\documentclass[journal]{IEEEtran}
\usepackage{graphicx}
\usepackage{cite}
\usepackage{picinpar}
\usepackage{amsmath}
\usepackage{url}
\usepackage{flushend}
\usepackage[latin1]{inputenc}
\usepackage{colortbl}
\usepackage{soul}
\usepackage{multirow}
\usepackage{pifont}
\usepackage{color}
\usepackage{alltt}
\usepackage[hidelinks]{hyperref}
\usepackage{enumerate}
\usepackage{siunitx}
\usepackage{breakurl}
\usepackage{epstopdf}
\usepackage{pbox}
\usepackage{makecell}
\usepackage{amssymb}
\usepackage{algorithm}
\usepackage{algorithmic}
\usepackage{amsthm,amsmath,amssymb}

\usepackage{mathrsfs}
\ifCLASSINFOpdf
\else
\fi
\begin{document}
\title{Brain-Inspired Modelling and Decision-making for Human-Like Autonomous Driving in Mixed Traffic Environment}

\author{Peng Hang, Yiran Zhang, and Chen Lv
       % <-this % stops a space
\thanks{This work was supported by the SUG-NAP Grant (No. M4082268.050) of Nanyang Technological University, Singapore.}% <-this % stops a space
\thanks{P. Hang, Y. Zhang, and C. Lv are with the School of Mechanical and Aerospace Engineering, Nanyang Technological University, Singapore 639798. (e-mail: \{peng.hang, yiran.zhang, lyuchen\}@ntu.edu.sg)}% <-this % stops a space

\thanks{Corresponding author: C. Lv}}


\markboth{ }%IEEE Transactions on Intelligent Transportation Systems
{Shell \MakeLowercase{\textit{et al.}}: }



\maketitle

\begin{abstract}

In this paper, a human-like driving framework is designed for autonomous vehicles (AVs), which aims to make AVs better integrate into the transportation ecology of human driving and eliminate the misunderstanding and incompatibility of human drivers to autonomous driving. Based on the analysis of the real world INTERACTION dataset, a driving aggressiveness estimation model is established with the fuzzy inference approach. Then, a human-like driving model, which integrates the brain emotional learning circuit model (BELCM) with the two-point preview model,  is designed. In the human-like lane-change decision-making algorithm, the cost function is designed comprehensively considering driving safety and travel efficiency. Based on the cost function and multi-constraint, the dynamic game algorithm is applied to modelling the interaction and decision making between AV and human driver. Additionally, to guarantee the lane-change safety of AVs, an artificial potential field model is built for collision risk assessment. Finally, the proposed algorithm is evaluated through human-in-the-loop experiments on a driving simulator, and the results demonstrated the feasibility and effectiveness of the proposed method.

\end{abstract}


\IEEEpeerreviewmaketitle

\section{Introduction}
\IEEEPARstart{A}{lthough} autonomous driving has made a great process in recent years, it can be believed that autonomous vehicles (AVs) and human-driven vehicles will coexist for a long time\cite{huang2021personalized}. It is a critical issue for AVs that human drivers feel safe and trustworthy in the interaction process with AVs, which requires AVs can merge into the transportation ecology of human driving and eliminate the misunderstanding and incompatibility of human drivers to autonomous driving\cite{hang2021cooperative}. Therefore, human-like driving and decision making become a research hotspot for AVs.

Various approaches and algorithms have been proposed for human-like driving and decision making\cite{gu2017human,zhang2020multi,yang2020preview}. In\cite{jiang2021personalized}, a personalized computational model is designed for human-like decision making under risk, which comprehensively considers regret effects, probability weighting effects and range effects. By analyzing the driving habits of different human drivers, a personalized human-like driver model is built, which includes a longitudinal driving behavior model and a lateral lane-change trajectory planning model\cite{yang2021personalized}. In\cite{cheng2020universal}, a concept of "staying within the safety zone" is proposed, and based on the concept, a universal control framework of human-like steering is designed, which can deal with multiple driving scenarios. In\cite{gu2017research}, a multi-point decision making strategy is proposed for human-like driving. To address the interaction and decision making between AV and human-driven vehicle at mixed-flow intersections, a logit model is combined with
Bayes' theorem to realize human-like interaction and autonomous driving\cite{zhou2019autonomous}. In\cite{rodrigues2018autonomous}, an adaptive tactical behavior planner (ATBP) is designed for AVs, which can plan human-like motion behaviors at a unsignalized roundabout.

Additionally, Markov Decision Process (MDP) is a common approach to simulate the human-like decision making process\cite{ulbrich2013probabilistic}. Based on MDP, a human-like longitudinal decision model is designed for AV to address the velocity planning at signalized intersections\cite{cheng2017human}. However, only longitudinal decision making is considered in this model. In\cite{xia2021human}, the lane-change intention of AV is described by a Hidden Markov Model (HMM) based intention recognizer. Based on HMM, a human-like lane-change intention understanding model is proposed for autonomous driving. In\cite{lin2019decision}, the partially observable Markov decision process (POMDP) is applied to the decision making for AVs at unsignalized intersections, which can conduct human-like driving behaviors. Based on MDP, a human-like driving model is constructed for AVs, which has three driving modes, i.e., leisure mode, normal mode and efficiency mode, to provide personalized driving demands for passengers\cite{guo2018toward}.

Leaning-based approaches have been widely applied to the human-like driving and decision making\cite{wu2021human,xu2020learning,sun2021brain}. Based on reinforcement learning (RL), a human-like longitudinal driver model is established for AVs\cite{xie2021modeling}. Besides, deep RL is used to design the human-like car-following model for AVs\cite{zhu2018human}. In\cite{zhang2018human}, double Q-leaning is adopted to realize human-like speed control. However, the above three models cannot deal with lateral decision making, e.g., lane-change. In\cite{li2019human}, general regression neural network (GRNN) is applied to the human-like lane-change trajectory generation. Based on the deep autoencoder (DAE) network and the XGBoost algorithm, a human-like lane-change decision making model is built\cite{gu2020novel}. In\cite{wang2018human}, Long Short Term Memory (LSTM) neural network and Conditional Random Field (CRF) model are combined together to construct the human-like maneuver decision model for AVs. Based on convolutional neural network, cognitive map and recurrent neural network, a brain-inspired cognitive model is established for human-like autonomous driving\cite{chen2017brain}. Although the leaning-based approaches can realize human-like and reasonable decision making, it requires a large amount of training data and the model accuracy is definitely affected by the quality of the training data set.

Although aforementioned models or approaches are in favor of human-like driving and decision making for AVs, few of them study the interaction between human driver and AV when dealing with driving conflicts. Game theory is capable of modelling the interactive and decision-making process between human driver and AV\cite{hang2020integrated}.
To address the lane-change issue of AVs, MDP is combined with game theory to realize human-like driving\cite{coskun2019receding}.
In\cite{smirnov2021game}, a game theory-based decision-making algorithm is proposed for AVs to realize human-like lane-change in congested urban intersections.
Besides, by estamiting the aggressiveness of surrounding drivers, a game theory-based lane-change model is built for AVs to mimic human behaviors\cite{yu2018human}. With the comprehensive consideration of safety and efficiency, a human-like decision making framework is proposed for AVs with the game theoretic approach and a single-point preview model\cite{hang2020human}.

In this paper, a novel human-like driving and decision-making framework is designed for AVs. The contributions of this paper are summarized as follows: (1) Based on the driving behavior analysis of human drivers with the INTERACTION dataset, the fuzzy inference approach is applied to the aggressiveness estimation of surrounding vehicles; (2) To realize human-like driving for AVs, the brain emotional learning circuit model (BELCM) is combined with the two-point preview model to construct the human-like driving model; (3) Based on the decision-making cost function that comprehensively considers driving safety and travel efficiency, the dynamic game theoretic approach is applied to the decision making of AVs. Finally, the effectiveness and feasibility of the proposed algorithm is verified via the human-in-the-loop experiments.

The remainder of this paper is organized as follows. The problem statement and the decision-making framework are presented in Section II. Section III is the modelling part including the vehicle model, aggressiveness estimation model and BELCM-based driving model. Then, the dynamic game theoretic approach is used to deal with the decision-making issue of AVs in Section IV. In Section V, the proposed algorithm is verified via human-in-the-loop experiments. Finally, Section VI presents the conclusion.

\section{Problem Description}
Although AVs can realize accurate control and decision making based on high-precision sensors and super computing platform, several avoidable accidents may happen between AVs and the human driven vehicles. The major cause is the misunderstanding between self-driving systems and human drivers. Human drivers usually follow some driving conventions, e.g, turning vehicles giving way to straight vehicles at unsignalized intersections. However, AVs will pass the intersection if the self-driving system thinks it is safe. The unconventional behaviors of AVs do not obey the driving logic of human drivers, which may lead to wrong decisions of human drivers and bring a collision risk. Therefore, it is necessary to consider the factors of human-like driving and decision making in the self-driving system, which can make AVs better integrate into the transportation ecology of human driving and eliminate the misunderstanding and incompatibility of human drivers to autonomous driving \cite{hang2020human}.

Since lane change is a common driving behavior and one of the most common causes of traffic accidents, especially in the highway condition, this paper mainly focuses on the study of human-like lane-change decision making for AVs. Fig. 1 shows a typical lane-change decision making scenario. Due to the slow speed of the black tractor, the red car must make decisions, i.e., slowing down and keeping the lane, or changing lanes. If choosing lane-change, the red car must fight with the blue truck. The blue truck can yield or fight. There exists an interaction and game between the two vehicles. To study the interaction and decision making between AVs and human driven vehicles. One of the two vehicles can be regarded as AV and another is human-driven vehicle. To describe the decision-making process, some names are defined for vehicles. Host vehicle (HV) is the vehicle that wants to make lane-change decision, i.e., the red car in Fig. 1. Neighbor conflict vehicle (NV) is the opponent for HV in the lane-change process. Leading vehicle (LV) is the vehicle in front of HV. For instance, the black tractor is the LV for the red car if choosing lane-keeping. After finishing the lane-change, the black bus is the LV for the red car. With the difference of HV's position, the role of NV and LV will change as well.

\begin{figure}[t]\centering
	\includegraphics[width=7.5cm]{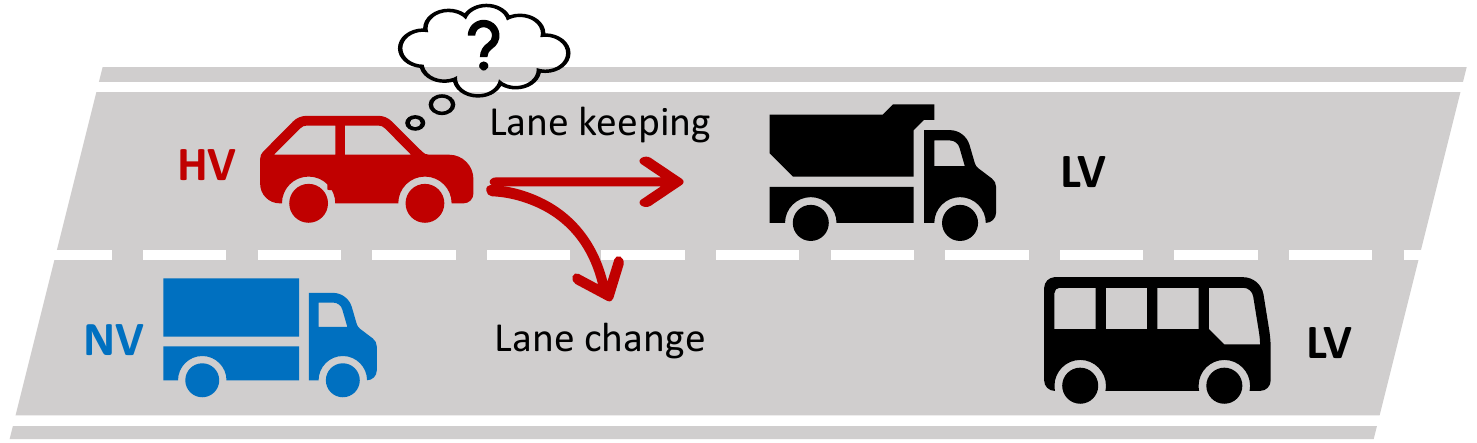}
	\caption{Lane-change decision making scenario.}\label{FIG_1}
\end{figure}

\begin{figure}[t]\centering
	\includegraphics[width=8.5cm]{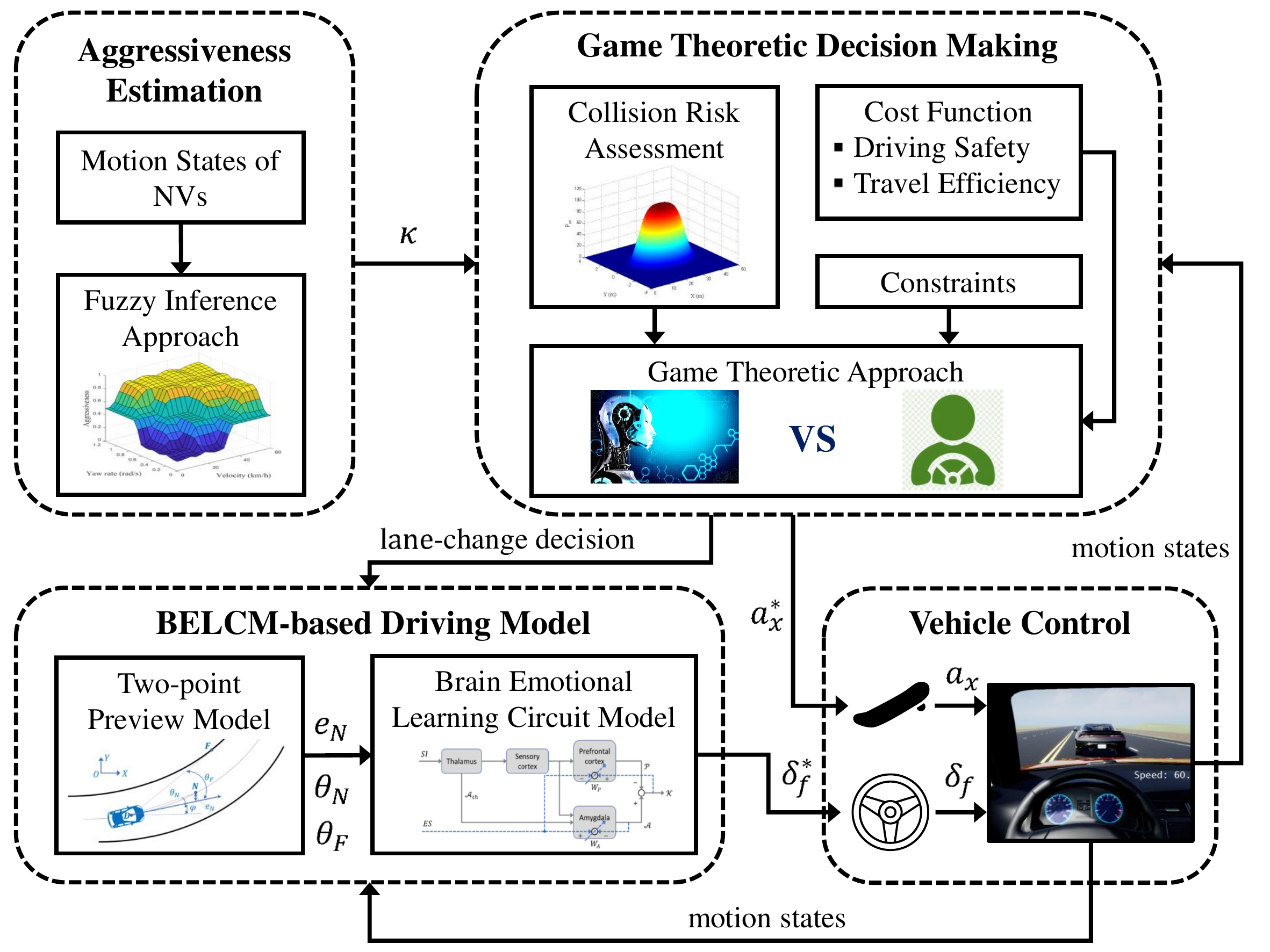}
	\caption{Human-like driving and decision-making framework for AVs.}\label{FIG_2}
\end{figure}

To deal with the above issue, a human-like driving and decision-making framework is designed, which is illustrated in Fig. 2. It mainly consists of three modules. Firstly, an aggressiveness estimation module is constructed based on the fuzzy inference approach. According to the motion states of NVs, the aggressiveness estimation results of NVs will be provided to the decision-making module of HV. In the human-like decision-making module, the cost function is designed comprehensively considering driving safety and travel efficiency. Based on the cost function and multi-constraint, the game theoretic approach is applied to the interaction and decision making between HV and NVs. Besides, a collision risk assessment algorithm is proposed to ensure the safety during the lane-change process. Then, the lane-change decision is provided to the human-like driving module, which is an integration of the two-point preview model and the brain emotional learning circuit model (BELCM). Finally, the ideal steering angle $\delta_f^*$ from the human-like driving module and the ideal acceleration $a_x^*$ from the human-like decision-making module are outputted to the vehicle control module to realize the motion control of AV.

\section{Modelling}
A simplified vehicle kinematic model is built for AVs. Then, according to the driving behavior analysis of human drivers, an aggressiveness estimation model is constructed for AVs using the fuzzy inference approach. Finally, a human-like driving model is built based on BELCM.

\subsection{Vehicle Model}
To simplify the computational model of decision making, the single-track model is adopted replacing the four-wheel vehicle model. The kinematic description of the single-track model is shown as follows \cite{hang2021decision}.
% Eq.
\begin{align}
\dot{x}(t)=F(x(t),u(t))
\end{align}
% Eq.
\begin{align}
F(x(t),u(t))=
&
\left[
\begin{array}{ccc}
a_x\\
v_x\tan\beta/b_r\\
v_x\cos\phi/\cos\beta\\
v_x\sin\phi/\cos\beta\\
\end{array}
\right]
\end{align}
% Eq.
\begin{align}
\beta=\arctan[b_r/(b_f+b_r)\tan\delta_f]
\end{align}
% Eq.
\begin{align}
\phi=\varphi+\beta
\end{align}
where $x=[v_x, \varphi, X, Y]^{T}$ and $u=[a_x, \delta_f]^{T}$ are the state vector and control vector. $v_x$, $\varphi$ and $\phi$ are the longitudinal velocity, yaw angle and heading angle. $X$ and $Y$ are the position coordinates of the center of gravity. $a_x$ and $\delta_f$ denote the longitudinal acceleration and the steering angle of the front wheel. $\beta$ denotes the sideslip angle. $b_f$ and $b_r$ denote the front and rear wheel bases of the vehicle.

\subsection{Aggressiveness Estimation Model}

\begin{figure}[t]\centering
	\includegraphics[width=8.5cm]{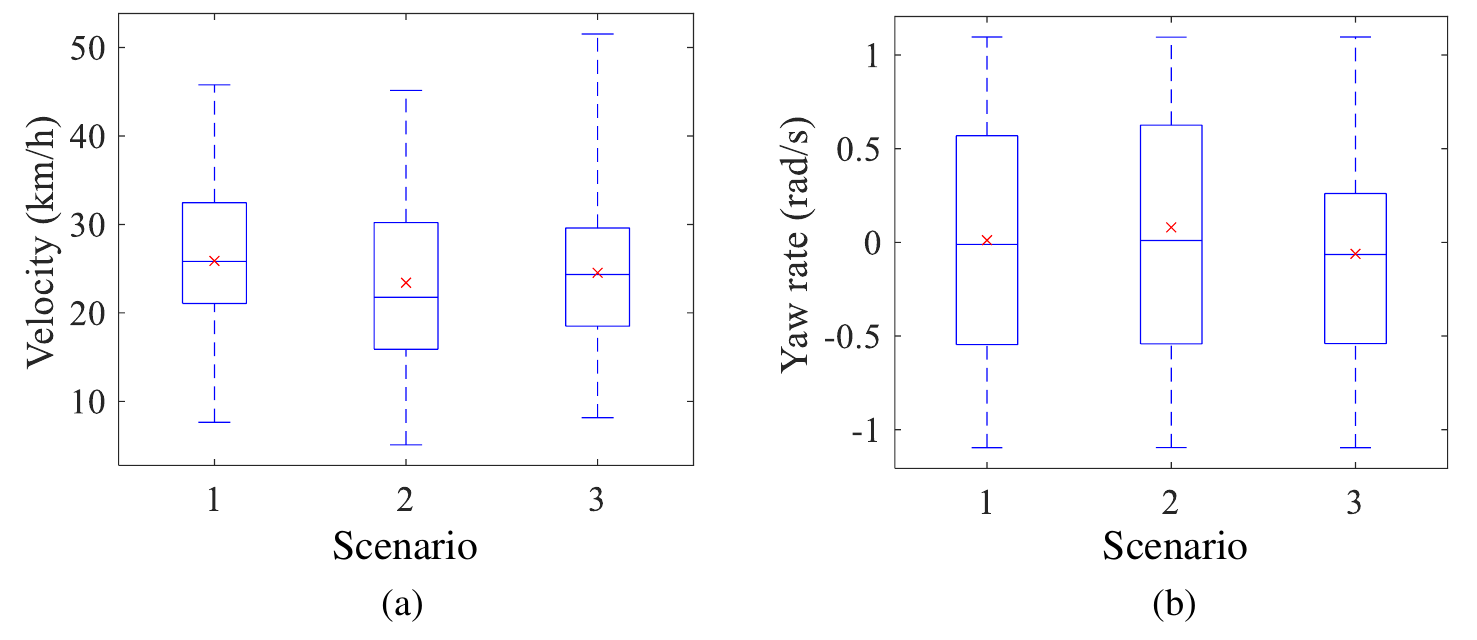}
	\caption{Driving behavior analysis of human drivers based on INTERACTION dataset: (a) Velocity; (b) Yaw rate.}\label{FIG_3}
\end{figure}

\begin{figure}[t]\centering
	\includegraphics[width=8.5cm]{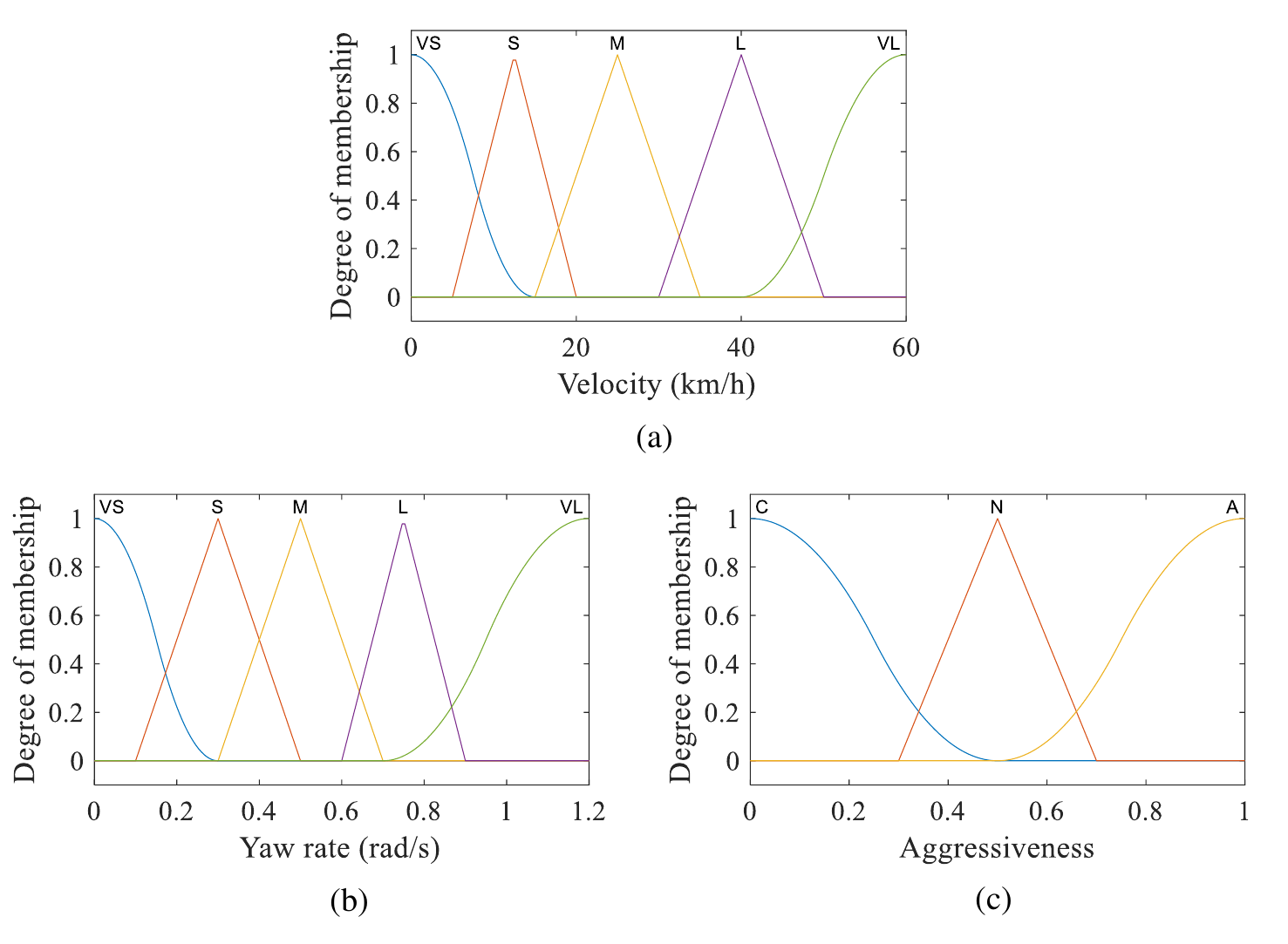}
	\caption{The membership functions: (a) Velocity; (b) Yaw rate; (c) Aggressiveness.}\label{FIG_4}
\end{figure}

\begin{figure}[t]\centering
	\includegraphics[width=6.5cm]{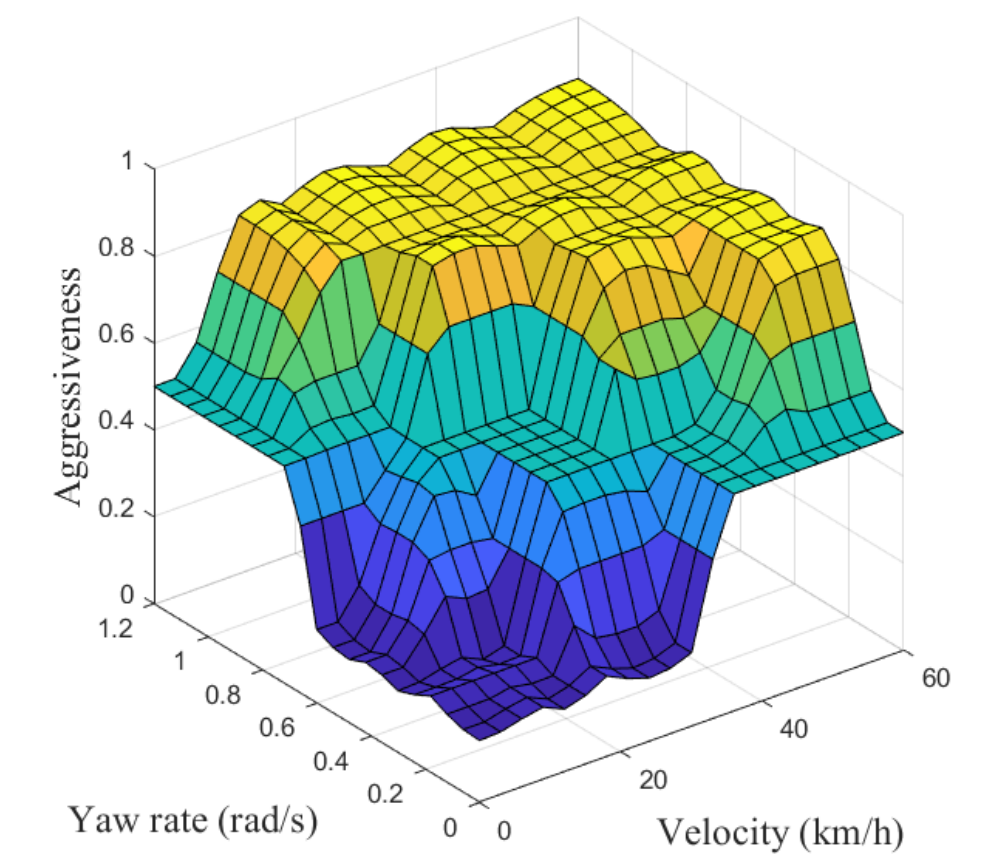}
	\caption{Aggressiveness MAP.}\label{FIG_5}
\end{figure}

\begin{table}[t]
	\renewcommand{\arraystretch}{1.2}
	\caption{Fuzzy Logic for Aggressiveness Estimation}
\setlength{\tabcolsep}{5.5 mm}
	\centering
	\label{table_1}
	%\centering
	\resizebox{\columnwidth}{!}{
		\begin{tabular}{c c c c c c}
			\hline\hline \\[-4mm]
			\multirow{2}{*}{Velocity} & \multicolumn{5}{c}{Yaw rate} \\
\cline{2-6} & \makecell [c] {VS} & \makecell [c] {S} & \makecell [c] {M} & \makecell [c] {L} & \makecell [c] {VL} \\
\hline
			\multicolumn{1}{c}{VS} & C & C & C & N & N \\
			\multicolumn{1}{c}{S} & C & C & N & N & A \\
            \multicolumn{1}{c}{M} & C & N & N & A & A \\
            \multicolumn{1}{c}{L} & N & N & A & A & A \\
            \multicolumn{1}{c}{VL} & N & A & A & A & A \\

			\hline\hline
		\end{tabular}
	}
\end{table}

To make AVs have different driving preferences like human drivers that is reflected by the aggressiveness index, the driving behaviors of human drivers are studied based on the INTERACTION dataset \cite{shi2010predicting}. The INTERACTION dataset contains naturalistic motions of various traffic participants. In this paper, the motion data of human-driven vehicles at three merging and lane-change scenarios are analyzed. The aggressiveness of human drivers is caused by two driving behaviors, i.e., accelerating or decelerating, and steering, mapping to the vehicle motion states vehicle velocity and yaw rate. As a result, the velocities and yaw rates of human-driven vehicles at three scenarios are illustrated in Fig. 3.

According to the distributions of velocity and yaw rate of vehicles, the fuzzy inference approach is used to estimate the aggressiveness of vehicles. The inputs of the fuzzy inference system are velocity and yaw rate, and finally, it outputs the aggressiveness. Three components make up the fuzzy inference system including fuzzification, the rule evaluation and the defuzzification \cite{huang2021toward}. The first step is fuzzification, which converts the continuous values of vehicle motion states into fuzzy values according to the membership functions. As Fig. 4 shows, Z-shaped and triangular membership functions are used. Velocity and yaw rate are blurred into very small (VS), small (S), middle(M), large(L), and very large(VL). Additionally, the fuzzy values of aggressiveness are conservative (C), normal (N) and aggressive (A). After finishing the fuzzification process of velocity and yaw rate, the fuzzy value of aggressiveness can be inferred based on the fuzzy logic rule in Table I. Once the fuzzy inference process is finished, the defuzzification will be conducted to obtain the real value of aggressiveness $\kappa$, $\kappa\in [0,1]$. Finally, the estimation of aggressiveness is done. Fig. 5 shows the map of aggressiveness with respect to velocity and yaw rate. It can be found that the larger the velocity and yaw rate, the higher the aggressiveness. If the velocity or yaw rate exceeds the maximum, $\kappa=1$.

\subsection{Brain Emotional Learning Circuit Model}

\begin{figure}[!t]\centering
	\includegraphics[width=9cm]{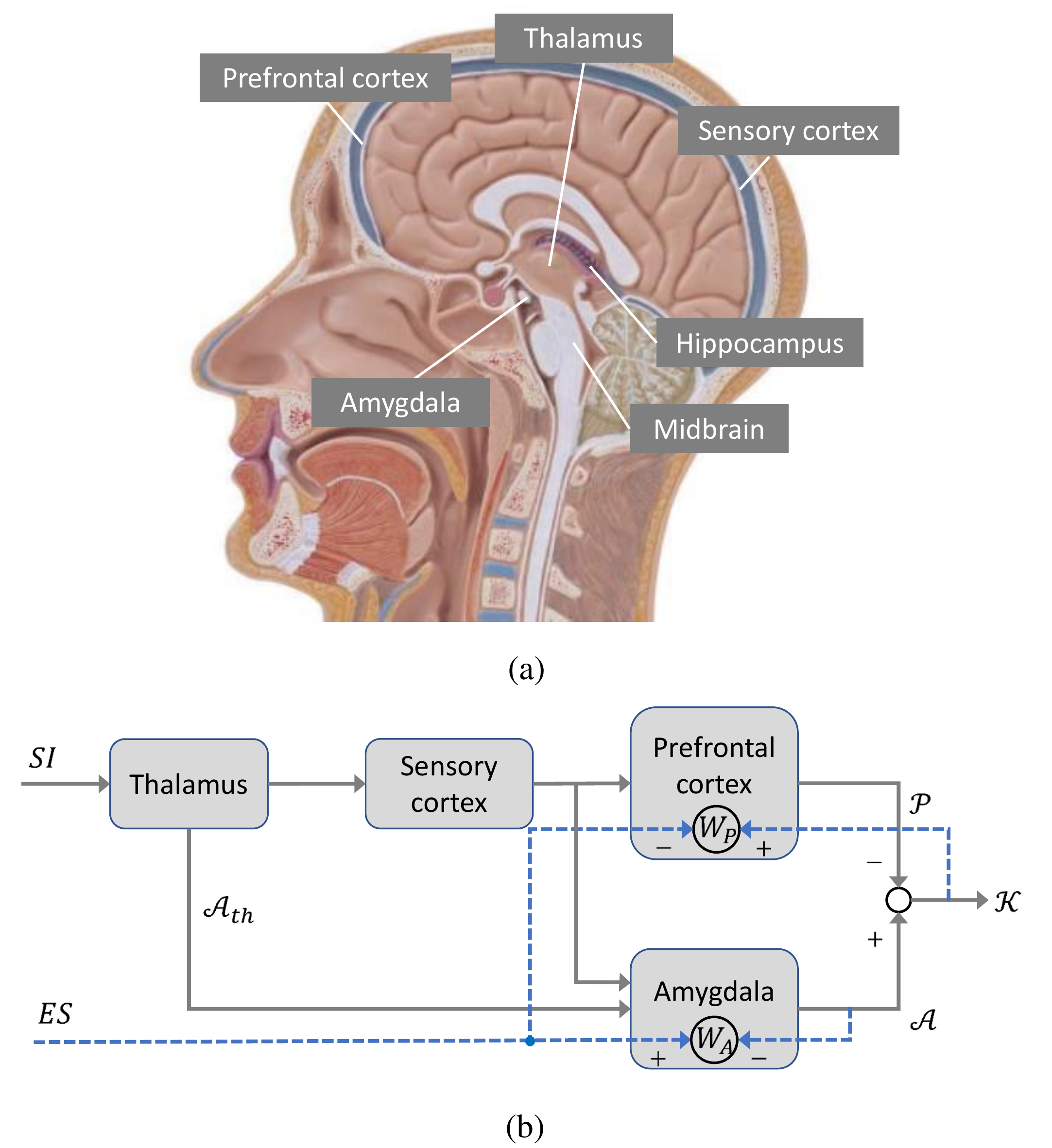}
	\caption{Brain emotional learning circuit model: (a)Brain structure; (b) Diagram of brain emotional learning circuit model\cite{2011Sensorless}.}\label{FIG_6}
\end{figure}

Brain-like computing has been widely studied in the field of artificial intelligence, which can simulate the operating mechanism of the human brain to establish a calculation model \cite{katayama2016wave}. As a result, the proposed model can realize human-like decision making and control in some intelligent applications. BELCM is a typical brain-like computing model proposed by Balkenius and Moren \cite{moren2001emotional}. Due to the simple structure and low computing load, it has been applied to the human-like controller design. In this section, BELCM is used to simulate the driving behaviors of human drivers.

The brain structure is illustrated in Fig. 6 (a), in which amygdala, prefrontal cortex, sensory cortex and thalamus are the key elements for brain emotional learning. The brain's emotional learning activities mainly occur in the amygdala, and the prefrontal cortex mainly focuses on monitoring the emotional learning process in the amygdala to avoid under-learning or over-learning. Fig. 6 (b) shows the structure diagram of BELCM. Firstly, the stimulus inputs (SIs) are processed by the thalamus, and the maximum SI is outputted to the amygdala, i.e.,
% Eq.
\begin{align}
\mathcal{A}_{th}=\max (SI_i),\quad i=1,2,\cdots,n
\end{align}

Then, other SIs are provided to the sensory cortex. After processing, the sensory signals are outputted to the prefrontal cortex and amygdala. Under the reference hint of the emotion signal (ES), the sensory signals in the amygdala start memory learning and finally output the learning signal $\mathcal{A}$. Additionally, the sensory signals in the prefrontal cortex will correct the emotional learning loop under the supervision of ES, and output the correction signal $\mathcal{P}$. Finally, the output of BELCM is expressed as
% Eq.
\begin{align}
\mathcal{K}=\mathcal{A}-\mathcal{P}
\end{align}

It can be found that the output $\mathcal{K}$ is determined by both the amygdala and prefrontal cortex. Therefore, it is necessary to study the learning process of amygdala and prefrontal cortex. There exits a corresponding node in the amygdala to each $SI_i$, and each amygdala node has a variable weighting coefficient $W_{A_i}$.
Then, the output of the amygdala can be written as
% Eq.
\begin{align}
\mathcal{A}=\mathcal{A}_{th}+\sum\limits_{i=1}^{n} \mathcal{A}_{i}
\end{align}
\begin{align}
\mathcal{A}_{i}=SI_{i}\cdot W_{A_i}
\end{align}

The change of $\mathcal{A}$ reflects the tracking process of the amygdala with respect to ES. The amygdala will adjust the weighting coefficient $W_{A_i}$ to decrease the tracking error of $\mathcal{A}$ and ES, which is a self-suggestion ability formed by the individual in long-term learning practice. From the analysis of biological characteristics, the adjustment rate of the weighting coefficient $W_{A_i}$ is defined by
% Eq.
\begin{align}
\Delta W_{A_i}=\alpha_A\cdot SI_i \cdot \max (0, ES-\mathcal{A}_{th}-\sum\limits_{i=1}^{n} \mathcal{A}_{i})
\end{align}
where $\alpha_A$ denotes the learning rate of the amygdala weighting coefficient, $\alpha_A\in (0,1)$ .

Similarly, there also exits a corresponding node in the prefrontal cortex to each $SI_i$, and each prefrontal cortex node has a variable weighting coefficient $W_{P_i}$.
Then, the output of the prefrontal cortex can be written as
% Eq.
\begin{align}
\mathcal{P}=\sum\limits_{i=1}^{n} \mathcal{P}_{i}
\end{align}
\begin{align}
\mathcal{P}_{i}=SI_{i}\cdot W_{P_i}
\end{align}

The change of $\mathcal{P}$ reflects the output correction of $\mathcal{A}$, which can be regarded as a counter-inhibition mechanism. The adjustment rate of the weighting coefficient $W_{P_i}$ is defined by
% Eq.
\begin{align}
\Delta W_{P_i}=\alpha_P\cdot SI_i \cdot (\sum\limits_{i=1}^{n} \mathcal{A}_{i}-\sum\limits_{i=1}^{n} \mathcal{P}_{i}-ES)
\end{align}
where $\alpha_P$ denotes the learning rate of the prefrontal cortex weighting coefficient, $\alpha_P\in (0,1)$ .

In general, after receiving SIs, the amygdala will conduct fast predictive learning under the reference hint of ES. Besides, the prefrontal cortex will correct the output of the amygdala under the supervision of ES. With the dynamically coordination of the two loops, i.e., memory learning and correction, BELCM can realize accurate tracking control.

\subsection{BELCM-based Human-Like Driving Model}

It has been studied that human drivers usually make decisions according two preview points, i.e., the near point and the far point \cite{lappi2014future}. The two-point preview driver model is illustrated in Fig. 7. The near point $N$ is located on the middle line of the lane. For a straight road, the far point $F$ is the vanishing point of sight. For a curve road, the far point $F$ is the tangent point 10-20 m in front of the vehicle on the inner edge of the road. The far point is used to estimate the road curvature and provide advance compensation control for smooth steering. The near point is previewed to compensate the position error and make the vehicle move in the middle of the lane. In Fig. 8, three key performance indexes for driver preview control are displayed, i.e., $\theta_N$, $e_N$ and $\theta_F$. $\theta_N$ and $e_N$ denote the preview angle and lateral error at the near point $N$. $\theta_F$ denotes the preview angle at the far point $F$. The three performance indexes can be perceived by human drivers and regarded as the feedback signals for vehicle decision making and control.

\begin{figure}[!t]\centering
	\includegraphics[width=8.5cm]{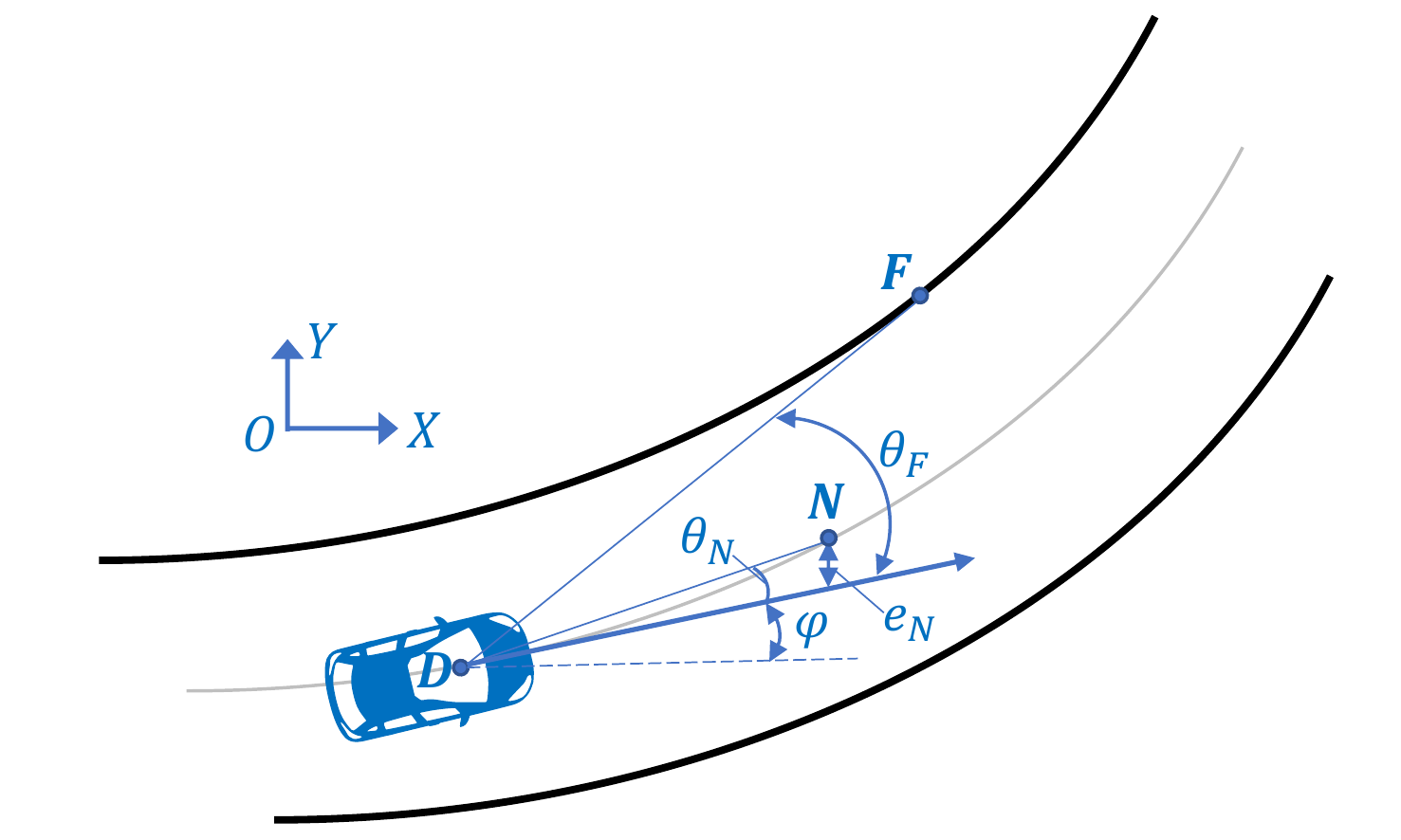}
	\caption{Two-point preview driver model.}\label{FIG_7}
\end{figure}

According to Fig. 7, $e_N$ can be derived as
% Eq.
\begin{align}
e_N=Y_N-Y-\tau_N v_x \varphi
\end{align}
where $Y_N$ and $Y$ are the coordinates of points $N$ and $D$ at Y-axis. The point $D$ is the driver position. $\tau_N$ is the preview time of the point $N$.

Additionally, $\theta_N$ and $\theta_F$ are expressed as
% Eq.
\begin{align}
\theta_N=\arctan \frac {Y_N-Y}{\tau_N v_x}-\varphi
\end{align}
% Eq.
\begin{align}
\theta_F=\arccos \frac{R_F}{R_D}
\end{align}
where $R_F$ and $R_D$ denote the curvature radiuses at points $F$ and $D$, respectively.

The control objectives of human drivers can be summarized as two points: safe path following, i.e., $e_N=0$ and $\theta_N=0$, smooth and comfort path following, i.e., $\dot{\theta}_F=0$. Therefore, $e_N$, $\theta_N$ and $\dot{\theta}_F$ are chosen as three SIs for the BELCM-based driver model, which is defined as follows.
% Eq.
\begin{align}
SI=[\eta_1 e_N, \eta_2 \theta_N, \eta_3 \dot{\theta}_F]^T
\end{align}
where $\eta_1$, $\eta_2$ and $\eta_3$ are the weighting coefficients of the three control performance indexes.

Furthermore, $ES$ is defined as
% Eq.
\begin{align}
ES=\varpi_1 e_N+\varpi_2 \theta_N+\varpi_3 \dot{\theta}_F+\varpi_4 \delta_f
\end{align}
where $\varpi_1$, $\varpi_2$, $\varpi_3$ and $\varpi_4$ are the weighting coefficients.

According to Fig. 7, the ideal control vector from the brain $\delta_f^*$ can be derived as follows.
% Eq.
\begin{align}
\begin{array}{lr}
\delta_f^*=\mathcal{K}=\mathcal{A}-\mathcal{P}\\
\quad=W_A
\left[
\begin{array}{ccc}
SI\\
\mathcal{A}_{th}\\
\end{array}
\right]
-W_P SI\\
\quad=\xi_1 e_N+\xi_2 \theta_N+\xi_3\dot{\theta}_F+W_{A_{th}}{A}_{th}\\
\end{array}
\end{align}
where
% Eq.
\begin{align}
W_A=[W_{A_1}, W_{A_2}, W_{A_3}, W_{A_{th}}]
\end{align}
% Eq.
\begin{align}
W_P=[W_{P_1}, W_{P_2}, W_{P_3}]
\end{align}
% Eq.
\begin{align}
\xi_i=\eta_i (W_{A_i}-W_{P_i}), \quad (i=1,2,3)
\end{align}

Due to the driver's physical delay from mental signal processing and muscular activation, the actual control output to the vehicle can be expressed as
% Eq.
\begin{align}
\delta_f=\frac {\delta_f^*}{1+\tau_d s}
\end{align}
where $\tau_d$ denotes the physical delay time.

Finally, the detailed workflow of the BELCM-based human-like driving model is displayed in Algorithm 1.

\begin{algorithm}[h]
\caption{Workflow of the BELCM-based human-like driving model.}
\begin{algorithmic}[1]
\STATE Parameter initialization: Define $\eta_1$, $\eta_2$, $\eta_3$, $\varpi_1$, $\varpi_2$, $\varpi_3$, $\varpi_4$, $W_A$, $W_P$, $\tau_N$, $\alpha_A$ and $\alpha_P$;
\STATE Calculate $SI$ and $ES$;
\STATE Calculate $\Delta W_{A_i}$ and $\Delta W_{P_i}$ to update $W_A$, $W_P$;
\STATE Calculate $\delta_f$ and output;
\STATE Go back to Step 2.
\end{algorithmic}
\end{algorithm}

\section{Game Theoretic Decision-Making Algorithm}
In this section, a collision risk assessment algorithm is proposed to guarantee the lane-change safety. Then, the decision-making cost function is designed, which comprehensively considers the driving safety and travel efficiency. Based on the decision-making cost function and constraints, dynamic game theory is applied to the lane-change decision making.

\subsection{Collision Risk Assessment during Lane-change Process}
To advance the safety of HV, artificial potential field (APF) approach is applied to the collision risk assessment.
The APF model for NV is constructed as follows \cite{hang2021path}.
% Eq.
\begin{align}
\Upsilon^{NV}(X, Y)=\lambda \exp \{-[\frac{(X-X^{NV})^2}{2\sigma_x^2}+\frac{(Y-Y^{NV})^2}{2\sigma_y^2}]^a\}
\end{align}
where $\Upsilon^{NV}(X, Y)$ denotes the distribution function of NV. $(X^{NV}, Y_{NV})$ is the position of NV. $a$ is the shape coefficient. $\sigma_x$ is associated with the vehicle length and driving aggressiveness, i.e., $\sigma_x=b_x e^\kappa L^{NV}$, in which $b_x$ is the proportional coefficient and $L^{NV}$ is the length of NV. $\sigma_y$ is related to the vehicle width, i.e., $\sigma_y=b_yW^{NV}$, in which $b_y$ is the proportional coefficient and $W^{NV}$ is the width of NV. $\lambda$ is a function of time to collision (TTC) and driving aggressiveness. Fig. 8 shows an instance of APF distribution for NV.
% Eq.
\begin{align}
\lambda= \frac{\lambda_0 e^\kappa}{(TTC+\epsilon)^2}
\end{align}
where $\lambda_0$ is a constant value and $\epsilon$ is a small value set to avoid zero denominator.

To realize safe lane change of HV, it is necessary to assess the collision risk between NV and HV during the lane-change process. An event-triggered mechanism (ETM) is defined according to the APF model.
% Eq.
\begin{align}
t_{k+1}\triangleq \inf \{t>t_k|\Upsilon>\Upsilon_{sf}\}
\end{align}
where $\Upsilon_{sf}$ denotes the safe APF value for HV to conduct safe lane change. Eq. 25 means that when $\Upsilon>\Upsilon_{sf}$, the trigger condition is satisfied. As a result, HV cannot execute lane-change order or continue the lane-change behavior. HV must choose lane-keeping, or return back to the original lane if lane-change is in progress.

\begin{figure}[t]\centering
	\includegraphics[width=8.5cm]{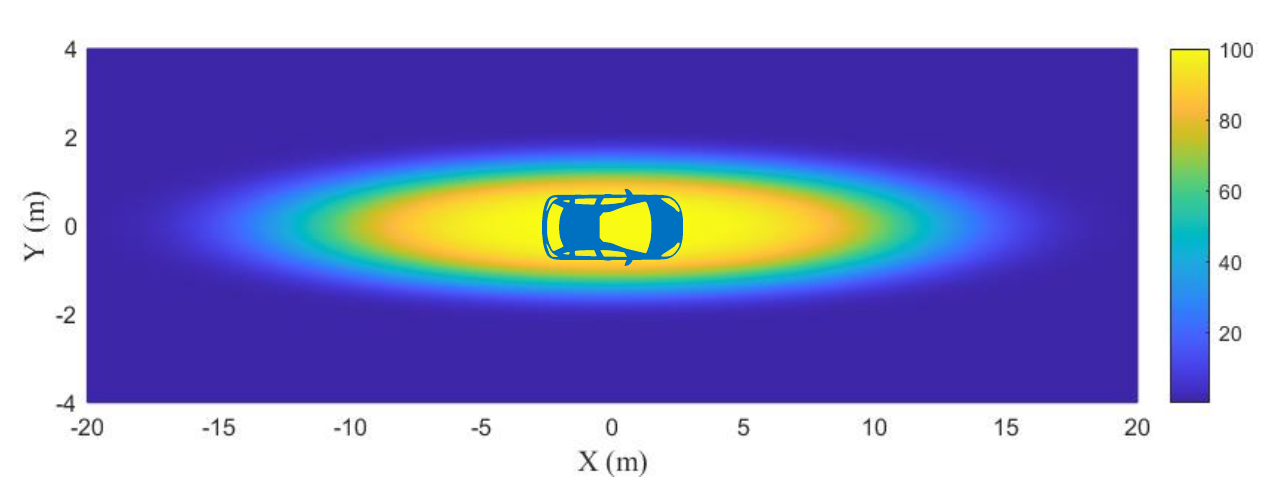}
	\caption{APF distribution of NV.}\label{FIG_8}
\end{figure}

\subsection{Construction of Cost Function}
In the decision-making cost function, both the driving safety and the travel efficiency are taken into account. Defining Vi as HV, the decision-making cost function $\Gamma^{i}$ for Vi is expressed as
% Eq.
\begin{align}
\Gamma^{i}=(1-\kappa^{i})\Gamma_s^{i}+\kappa^{i}\vartheta^i\Gamma_e^{i}
\end{align}
where $\Gamma_s^{i}$ and $\Gamma_e^{i}$ denote the costs of driving safety and travel efficiency. The weights of the two driving performance indexes are associated with the aggressiveness $\kappa^{i}$, i.e., a negative correlation between $\kappa^{i}$ and $\Gamma_s^{i}$, and a positive correlation for $\kappa^{i}$ and $\Gamma_e^{i}$. $\vartheta^i$ is a normalization parameter.

The cost function of driving safety  $\Gamma_s^{i}$ consists of the longitudinal safety cost and the lateral safety cost.
% Eq.
\begin{align}
\Gamma_s^{i}=k_{s-log}^{i}(1-(\alpha^i)^2) \Gamma_{s-log}^{i}+k_{s-lat}^{i} (\alpha^i)^2 \Gamma_{s-lat}^{i}
\end{align}
where $\Gamma_{s-log}^{i}$ and $\Gamma_{s-lat}^{i}$ denote the costs of the longitudinal and lateral safety, respectively. $k_{s-log}^{i}$, $k_{s-lat}^{i}$ are the weights of the two safety indexes. $\alpha^i \in \{-1, 0, 1\}$ denotes the lane-change decision making of Vi, i.e., left lane change ($\alpha^i=-1$), lane keeping ($\alpha^i=0$), and right lane change ($\alpha^i=1$).

$\Gamma_{s-log}^{i}$ reflects the longitudinal safety performance of Vi in the decision making process, which is a function of the relative longitudinal gap $\Delta s_{x,log}^i$ and relative longitudinal velocity $\Delta v_{x,log}^i$ between Vi and its LV.
% Eq.
\begin{align}
\Gamma_{s-log}^{i}=\omega_{v-log}^i\zeta_{v-log}^i(\Delta v_{x,log}^i)^2+\omega_{s-log}^i/[(\Delta s_{x,log}^i)^2+\epsilon]
\end{align}
% Eq.
\begin{align}
\Delta v_{x,log}^i=v_{x}^{LV}-v_{x}^{i}
\end{align}
% Eq.
\begin{align}
\Delta s_{x,log}^i=\sqrt {(X^{LV}-X^i)^2+(Y^{LV}-Y^i)^2}
\end{align}
% Eq.
\begin{align}
\zeta_{v-log}^i=
&
\left\{
\begin{array}{lr}
0,\quad \Delta v_{x,log}^i\geq0\\
1,\quad \Delta v_{x,log}^i<0\\
\end{array}
\right.
\end{align}
where $v_{x}^{LV}$ and $v_{x}^{i}$ denote the longitudinal velocities of LV and Vi. $(X^{LV}, Y^{LV})$ and $(X^{i}, Y^{i})$ are the position coordinates of LV and AVi. $\omega_{v-log}^i$ and $\omega_{s-log}^i$ are the weighting coefficients.

The cost of lateral safety $\Gamma_{s-lat}^{i}$ is proposed to address the merging conflict of Vi and NV, which is a function of the relative longitudinal gap $\Delta s_{x,lat}^i$ and relative longitudinal velocity $\Delta v_{x,lat}^i$ between Vi and NV.
% Eq.
\begin{align}
\Gamma_{s-lat}^{i}=\omega_{v-lat}^i\zeta_{v-lat}^i(\Delta v_{x,lat}^i)^2+\omega_{s-lat}^i/[(\Delta s_{x,lat}^i)^2+\epsilon]
\end{align}
% Eq.
\begin{align}
\Delta v_{x,lat}^i=v_{x}^{i}-v_{x}^{NV}
\end{align}
% Eq.
\begin{align}
\Delta s_{x,lat}^i=\sqrt {(X^{NV}-X^i)^2+(Y^{NV}-Y^i)^2}
\end{align}
% Eq.
\begin{align}
\zeta_{v-lat}^i=
&
\left\{
\begin{array}{lr}
0,\quad \Delta v_{x,lat}^i\geq0\\
1,\quad \Delta v_{x,lat}^i<0\\
\end{array}
\right.
\end{align}
where $v_{x}^{NV}$ denotes the longitudinal velocity of NV. $(X^{NV}, Y^{NV})$ is the position coordinate of NV. $\omega_{v-lat}^i$ and $\omega_{s-lat}^i$ are the weighting coefficients.

Besides, the cost function of passing efficiency $\Gamma_{e}^{i}$ for Vi is defined by the velocity error between $v_{x}^{i}$ and the maximum velocity $v_x^{\max}$.
% Eq.
\begin{align}
\Gamma_{e}^{i}=k_e^i(v_x^{i}-v_x^{\max})^2
\end{align}
where $k_e^i$ is the weighting coefficient.

\subsection{Decision Making with Dynamic Game}
If Vi (AV) and Vj (human driver) are regarded as two players, the lane-change decision making issue can be transformed into a dynamic game. Both Vi and Vj aim to minimize its own decision-making cost function. The decision-making cost functions of Vi and Vj are denoted by $\Gamma^i$ and $\Gamma^j$. The decision-making vectors of Vi and Vj are denoted by $u^i$ and $u^j$, $u^{i}=[\alpha^i, a_x^i]^T \in U^i$, $u^{j}=[\alpha^j, a_x^j]^T \in U^j$. $U^i$ and $U^j$ are the strategy set of Vi and Vj, respectively.

If the following conditions hold, $\{u^{i\ast},u^{j\ast}\}$ can be regarded as a Nash equilibrium for the dynamic game\cite{hang2020integrated}.
% Eq.
\begin{align}
\begin{array}{lr}
\Gamma^i (u^{i},u^{j\ast})\geq \Gamma^i (u^{i\ast},u^{j\ast})\\
\Gamma^j (u^{i\ast},u^{j})\geq \Gamma^j (u^{i\ast},u^{j\ast})\\
\end{array}
\end{align}
which indicates that no player can decrease its cost function by single-mindedly changing its strategy, as long as the other player sticks to the equilibrium strategy.

Furthermore, $\{u^{i\ast},u^{j\ast}\}$ can be derived by
% Eq.
\begin{align}
\begin{array}{lr}
u^{i\ast}=\arg\underset{u^i}{\min} \Gamma^i(u^{i}, u^{j\ast})\\
u^{j\ast}=\arg\underset{u^j}{\min} \Gamma^i(u^{i\ast}, u^{j})\\
\end{array}
\end{align}
subject to % Eq.
\begin{align}
\begin{array}{lr}
\dot{x}(t)=f(t, x, u^{i}, u^{j\ast})\\
\dot{x}(t)=f(t, x, u^{i\ast},  u^{j})\\
x(0)=x_0,\quad u^i\in U^i, \quad u^j\in U^j
\end{array}
\end{align}

The existence of Nash equilibrium can be guaranteed by fixed point theorem, and heuristic dynamic programming (HDP) is used to solve the Nash equilibrium\cite{venayagamoorthy2002comparison}.

\section{Algorithm Validation via Human-in-the-loop Experiments}
To evaluate the performance of the proposed human-like driving and decision-making framework for AVs, the human-in-the-loop experiments are conducted with the driving simulator.

\begin{figure}[t]\centering
	\includegraphics[width=7.5cm]{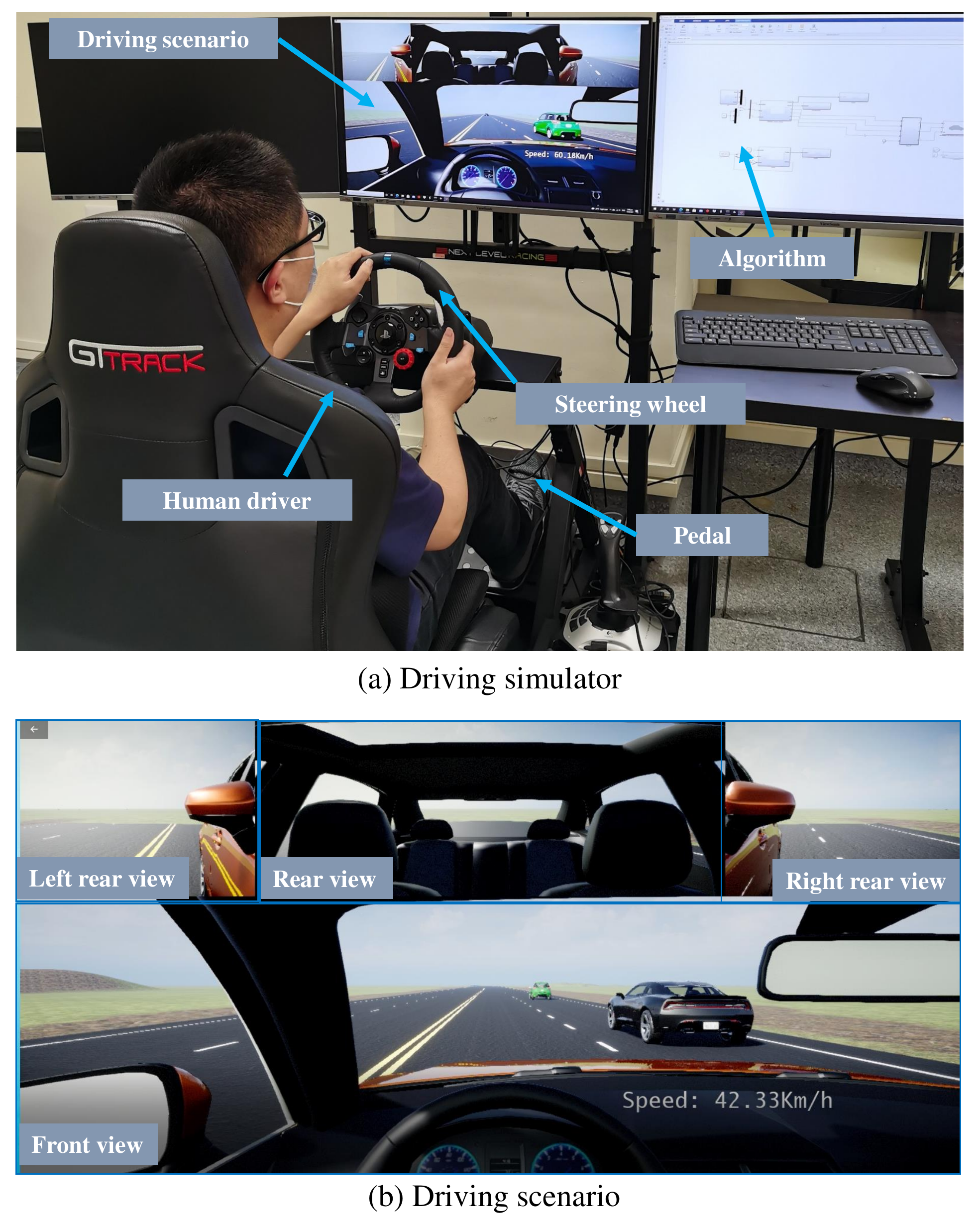}
	\caption{Human-in-the-loop experimental platform.}\label{FIG_9}
\end{figure}

\begin{figure}[t]\centering
	\includegraphics[width=7cm]{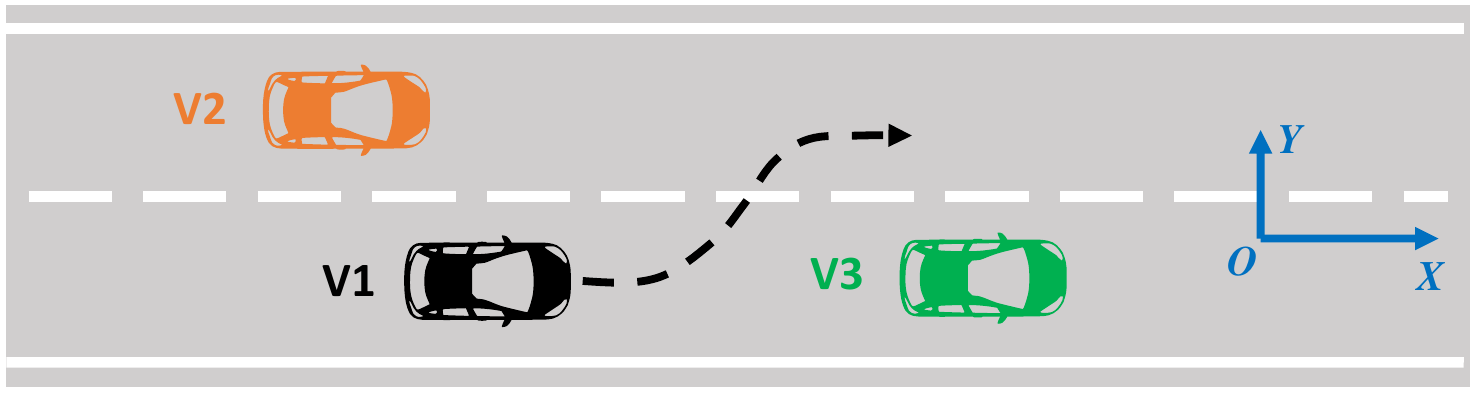}
	\caption{Lane-change decision making scenario.}\label{FIG_10}
\end{figure}

\begin{table}[t]
	\renewcommand{\arraystretch}{1.2}
	\caption{Parameters for Driving Behavior Analysis of Human Driver}
\setlength{\tabcolsep}{1 mm}
	\centering
	\label{table_1}
	%\centering
	\resizebox{\columnwidth}{!}{
		\begin{tabular}{c c c c c c c c c c}
			\hline\hline \\[-4mm]
			\multirow{2}{*}{Parameter} & \multicolumn{9}{c}{Case} \\
\cline{2-10} & \makecell [c] {1} & \makecell [c] {2} & \makecell [c] {3} & \makecell [c] {4} & \makecell [c] {5} & \makecell [c] {6} & \makecell [c] {7} & \makecell [c] {8} & \makecell [c] {9} \\
\hline
			\multicolumn{1}{c}{$X$ of V2 / (m)} & -130 & -130 & -130 & -120 & -120 & -120 & -110 & -110 & -110\\
			\multicolumn{1}{c}{$v_x$ of V2 / (m/s)} & 8 & 10 & 12 & 8 & 10 & 12 & 8 & 10 & 12\\

			\hline\hline
		\end{tabular}
	}
\end{table}
\subsection{Human-in-the-loop Experimental Platform}
The human-in-the-loop experimental platform is shown in Fig. 9, in which the driving simulator mainly consists of a computer equipped with a 11th Gen Intel Core i9 CUP and an NVIDIA GTX 3080 Super GPU, three joint head-up monitors, and the Logitech G29 steering wheel suit. Besides, the driving scenario is constructed based on the Unreal Engine and Simulink, which is shown in Fig. 9 (b). Four views are considered for the human driver, i.e., front view, left rear view, rear view, and right rear view.

\subsection{Driving Behavior Analysis of Human Driver and Human-Like Parameter Setting for AV}

\begin{table*}[!t]
	\renewcommand{\arraystretch}{1.3}
	\caption{Lane-change behavior analysis of human drivers}
\setlength{\tabcolsep}{2 mm}
	\centering
	\label{table_3}
	%\centering
	\resizebox{\textwidth}{15mm}{
		\begin{tabular}{c c c | c c | c c | c c | c c | c c | c c | c c | c c c c c c c c}
			\hline\hline \\[-4mm]
			\multirow{2}{*}{Test results} & \multicolumn{2}{c|}{Case 1} & \multicolumn{2}{c|}{Case 2} & \multicolumn{2}{c|}{Case 3} & \multicolumn{2}{c|}{Case 4} & \multicolumn{2}{c|}{Case 5} & \multicolumn{2}{c|}{Case 6} & \multicolumn{2}{c|}{Case 7} & \multicolumn{2}{c|}{Case 8} & \multicolumn{2}{c}{Case 9} \\
\cline{2-19} & \makecell [c] {HD-A} & \makecell [c] {HD-B} & \makecell [c] {HD-A} & \makecell [c] {HD-B} & \makecell [c] {HD-A} & \makecell [c] {HD-B} & \makecell [c] {HD-A} & \makecell [c] {HD-B} & \makecell [c] {HD-A} & \makecell [c] {HD-B} & \makecell [c] {HD-A} & \makecell [c] {HD-B} & \makecell [c] {HD-A} & \makecell [c] {HD-B} & \makecell [c] {HD-A} & \makecell [c] {HD-B} & \makecell [c] {HD-A} & \makecell [c] {HD-B}\\
\hline
    \multicolumn{1}{l}{TTL Avg/ (s)} & 3.33 & 3.06 & 3.18 & 3.11 & 3.26 & 2.97 & 3.32 & 3.02 & 3.32 & 2.93 & 3.30 & 2.87 & 3.35 & 2.78 & 3.21 & 2.64 & 3.12 & 2.64\\
    \multicolumn{1}{l}{LCT Avg/ (s)} & 7.17 & 5.18 & 6.82 & 4.96 & 6.08 & 4.88 & 7.07 & 4.92 & 6.47 & 4.90 & 6.32 & 4.88 & 6.39 & 4.64 & 6.38 & 4.52 & 6.14 & 4.42\\
\multicolumn{1}{l}{SY Avg / (m)} & 1.11 & 0.96 & 1.22 & 1.07 & 1.08 & 0.94 & 1.07 & 0.86 & 0.98 & 1.01 & 0.97 & 1.01 & 1.02 & 0.96 & 1.08 & 0.99 & 0.95 & 0.98\\
              \multicolumn{1}{l}{$v_x$ Max / (m/s)} & 12.45 & 16.73 & 14.16 & 16.70 & 14.38 & 16.81 & 12.67 & 16.76 & 13.50 & 16.77 & 14.67 & 16.86 & 14.65 & 16.97 & 15.66 & 17.08 & 16.11 & 17.21\\
             \multicolumn{1}{l}{$v_x$  Avg / (m/s)} & 12.15 & 14.81 & 12.90 & 15.08 & 12.94 & 15.52 & 12.18 & 15.28 & 12.91 & 15.49 & 13.29 & 15.77 & 13.51 & 15.98 & 14.18 & 16.04 & 14.92 & 16.27\\

			\hline\hline
		\end{tabular}
	}
\end{table*}

This test case aims to study the driving behavior and lane-change decision making of human drivers and apply them to human-like parameter setting for AV. As Fig. 10 shows, V1 is controlled by human driver to conduct the lane-change behavior, V2 and V3 are set to be moving at constant speeds. The initial position coordinates of V1 and V3 are set as (-100, -3) and (-65, -3), respectively. The initial velocities of V1 and V3 are set as 12 m/s and 5 m/s, respectively. The $Y$ coordinate position of V2 is 1. By controlling the relative distance and velocity between V1 and V2, it yields 9 cases in Table II. Tow human drivers, i.e., human driver A (HD-A, 3 years driving experience) and human driver B (HD-B, 10 years driving experience), are considered in the test. In each case, ten repeated tests are conducted.

\begin{figure}[t]\centering
	\includegraphics[width=8.5cm]{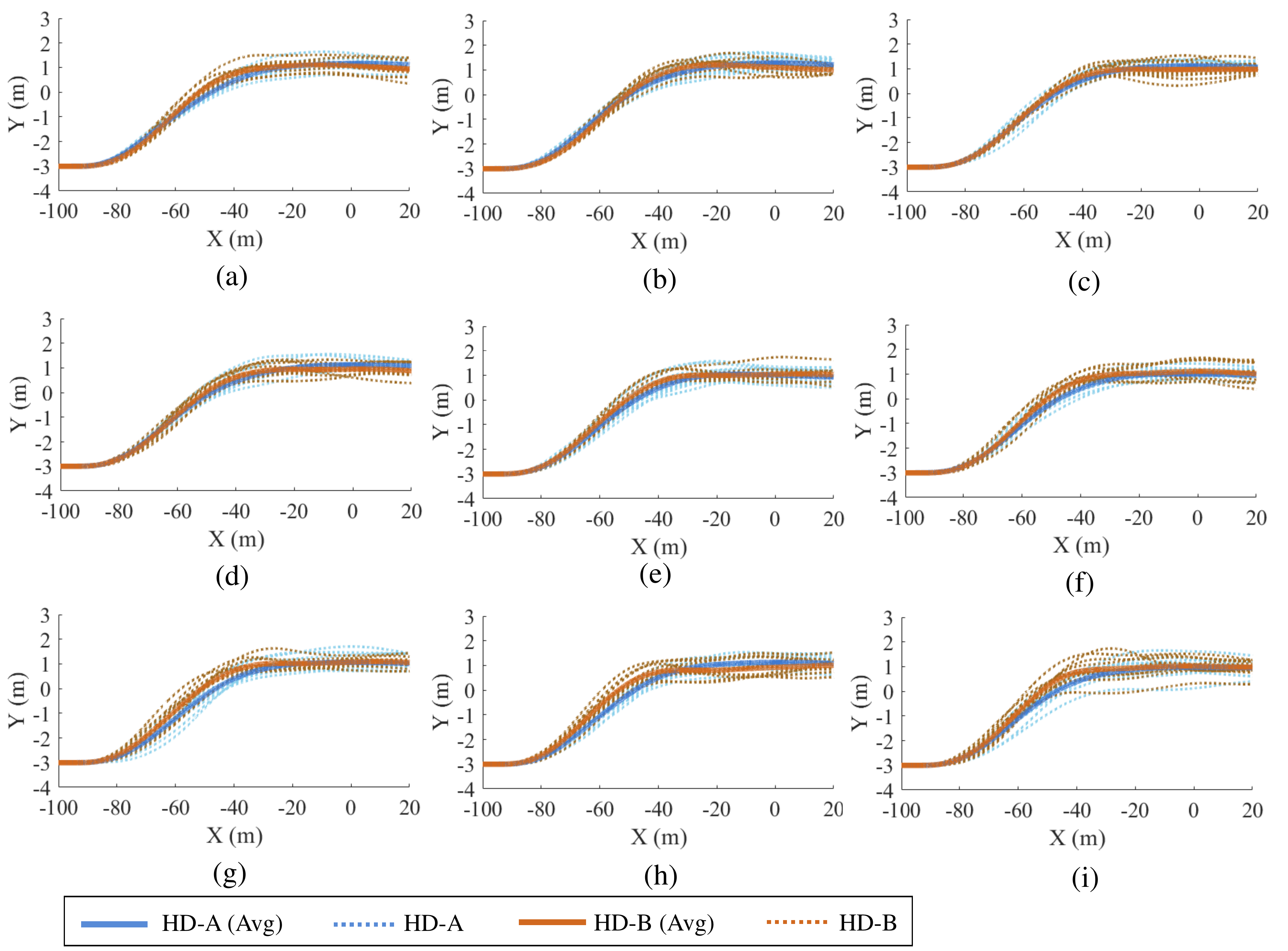}
	\caption{Lane-change trajectories of human drivers: (a) Case 1; (b) Case 2; (c) Case 3; (d) Case 4; (e) Case 5; (f) Case 6; (g) Case 7; (h) Case 8; (i) Case 9.}\label{FIG_11}
\end{figure}

\begin{figure}[t]\centering
	\includegraphics[width=8.5cm]{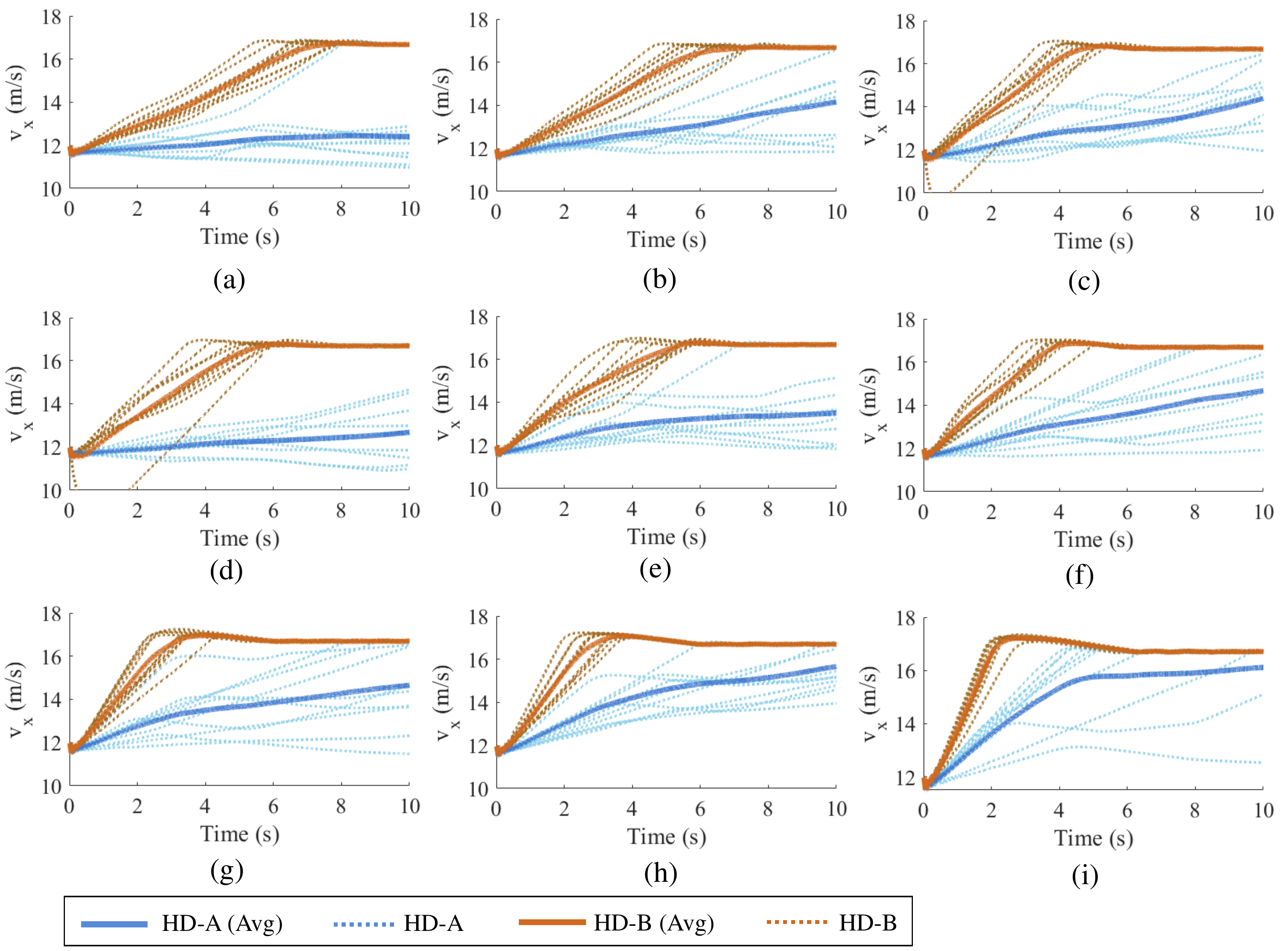}
	\caption{Lane-change velocities of human drivers: (a) Case 1; (b) Case 2; (c) Case 3; (d) Case 4; (e) Case 5; (f) Case 6; (g) Case 7; (h) Case 8; (i) Case 9.}\label{FIG_12}
\end{figure}

The lane-change trajectories and velocities of HD-A and HD-B are illustrated in Figs. 11 and 12, respectively. For each case, the lane-change trajectories have a high degree of coincidence in the first half, which indicates HD-A has similar lane-change start time and steering behaviors for the five tests in each case. Due to the difference of lane-change ending position, there exit some steady position errors within the acceptable range. Besides, the velocity has similar trend in each case. As the test case become more and more severe, HD-A increases the driving velocity to advance the safety during the lane-change process. After analyzing from the test results of trajectories and velocities for HD-B, the similar conclusions can be drawn. The difference from HD-A is that the lane-change ending time of HD-B is earlier than HD-A. Besides, HD-B has larger velocity than HD-A in all cases, which means HD-B is more aggressive than HD-A.

The detailed analysis of human driving behaviors is displayed in Table III. Three evaluation indexes of lane-change behavior are proposed, i.e., time to lane line (TTL), lane-change time (LCT), and steady $Y$ (SY) after lane-change. For TTL, the lane line is the crossed lane line during the lane-change process. TTL and LCT
reflect the lane-change speed, and SY reflects the path-tracking performance in the steady state, which are all related to the lateral driving and decision-making behaviors of human drivers. Besides, $v_x$ is used to evaluate the longitudinal driving and decision-making behaviors of human drivers during the lane-change process. Some findings are concluded as follows.

The two human drivers have different driving styles and characteristics to conduct the lane-change maneuver. From the test results of TTL and LCT, we can find that HD-B has faster lane-change speed than HD-A, especially in the most extreme case. From the test result of SY, it can be found that both HD-A and HD-B have small path tracking error (The target path is the lane centerline, i.e., $Y=1$. ) in most cases. Besides, it can also be found that HD-B has larger $v_x$ than HD-A when conducting lane-change, which also reflects stronger driving aggressiveness of HD-B.

Different initial states and positions of vehicles lead to different driving behaviors and lane-change decisions. With the reduction of initial relative distance between V1 and V2, both TTL and LCT decrease, and $v_x$ increases, which indicates human drivers want to change lanes as quick as possible to guarantee a safe distance from the obstacle vehicle. Besides, with the rise of V2's initial velocity, the similar conclusion can be drawn. Human drivers must accelerate and change lanes quickly to reduce the collision risk in the lane-change process. The decision-making differences between human drivers are highlighted in extreme and emergency conditions.

According to the test and analysis results of the two human drivers, the human-like parameter setting of AV for driving and decision making is realized, shown in Table IV. As a result, by simulating HD-A and HD-B, two kinds of human-like model (HLM-A and HLM-B) for driving and decision making are designed.

\begin{table}[t]
	\renewcommand{\arraystretch}{1.2}
	\caption{Parameters for Human-Like Driving and Decision-Making Models}
\setlength{\tabcolsep}{2 mm}
	\centering
	\label{table_1}
	%\centering
	\resizebox{\columnwidth}{!}{
		\begin{tabular}{c c c | c c c c }
			\hline\hline \\[-4mm]
\multicolumn{1}{c}{Parameter} & \multicolumn{1}{c}{HLM-A} & \multicolumn{1}{c|}{HLM-B} & \multicolumn{1}{c}{Parameter} & \multicolumn{1}{c}{HLM-A} & \multicolumn{1}{c}{HLM-B}\\
\hline
			$\eta_1$ & $2\times10^{-6}$ & $10^{-5}$ & $\varpi_1$ & 5 & 1 \\
			$\eta_2$ & $5\times10^{-5}$ & $10^{-4}$ & $\varpi_2$ & $3\times10^{-3}$ & $10^{-3}$ \\
            $\eta_3$ & 0.3 & 0.1 & $\varpi_3$ & $6\times10^{-3}$ & $10^{-3}$ \\
            $\alpha_A$ & 0.2 & 0.1 & $\varpi_4$ & 1500 & 1000 \\
            $\alpha_B$ & 0.2 & 0.1 & $\omega_{v-log}$ & 3 & 2 \\
            $k_{s-log}$ & 10 & 6 & $\omega_{s-log}$ & 8 & 5 \\
            $k_{s-lat}$ & 80 & 50 & $\omega_{v-lat}$ & 3 & 2 \\
            $k_e$ & 8 & 15 & $\omega_{s-lat}$ & 8 & 5 \\
			\hline\hline
		\end{tabular}
	}
\end{table}

\subsection{Interaction and Decision Making Experiments between Human Driver and AV}

\begin{table*}[!t]
	\renewcommand{\arraystretch}{1.3}
	\caption{Test result analysis in the interaction and decision making experiments between human driver and AV}
\setlength{\tabcolsep}{3 mm}
	\centering
	\label{table_5}
	%\centering
	\resizebox{\textwidth}{38mm}{
		\begin{tabular}{c l | c c c c | c c c c |c c c c}
			\hline\hline \\[-4mm]
			\multicolumn{2}{c|}{} & \multicolumn{4}{c|}{Case 1} & \multicolumn{4}{c|}{Case 2} & \multicolumn{4}{c}{Case 3}\\
\cline{3-14} \multicolumn{2}{c|}{vs} & \makecell [c] {HD-A} & \makecell [c] {HLM-A} & \makecell [c] {HD-B} & \makecell [c] {HLM-B} & \makecell [c] {HD-A} & \makecell [c] {HLM-A} & \makecell [c] {HD-B} & \makecell [c] {HLM-B} & \makecell [c] {HD-A} & \makecell [c] {HLM-A} & \makecell [c] {HD-B} & \makecell [c] {HLM-B}\\
\hline
         \multirow{6}{*}{Competitor-1} & TTL Avg / (s) & 4.33 & 4.20 & 3.42 & 3.45 & 3.28 & 3.19 & 3.19 & 3.16 & 7.91 & 7.48 & 3.07 & 3.02\\
                             ~& LCT Avg / (s) & 6.57 & 6.60 & 5.88 & 5.38 & 5.10 & 5.15 & 5.08 & 5.12 & -- & -- & 5.21 & 5.18\\
                              ~& SY Avg / (m) & 0.61 & 1.00 & 0.78 & 1.00 & 0.76 & 1.00 & 0.64 & 1.00 & -- & -- & 0.77 & 1.00\\
                         ~& $v_x$ Max / (m/s) & 16.98 & 16.66 & 16.82 & 16.66 & 17.11 & 16.67 & 17.08 & 16.67 & 17.22 & 16.61 & 17.23 & 16.68\\
                        ~& $v_x$  Avg / (m/s) & 15.06 & 15.30 & 15.55 & 15.62 & 15.95 & 15.93 & 15.94 & 15.94 & 12.29 & 12.29 & 16.20 & 16.12\\
                             ~& $\kappa$  Avg & 0.53 & 0.53 & 0.52 & 0.54 & 0.54 & 0.54 & 0.53 & 0.51 & 0.46 & 0.47 & 0.55 & 0.56\\
\hline
         \multirow{6}{*}{Competitor-2} & TTL Avg / (s) & 4.21 & 4.00 & 3.19 & 3.22 & 4.33 & 4.03 & 3.33 & 3.26 & 7.95 & 7.44 & 3.37 & 3.10\\
                             ~& LCT Avg / (s) & 6.69 & 6.63 & 6.35 & 5.12 & 6.80 & 6.88 & 5.67 & 5.37 & -- & -- & 6.09 & 5.35\\
                              ~& SY Avg / (m) & 1.39 & 1.00 & 0.79 & 1.00 & 0.99 & 1.00 & 0.84 & 1.00 & -- & -- & 0.96 & 1.00\\
                         ~& $v_x$ Max / (m/s) & 17.19 & 16.66 & 16.72 & 16.66 & 17.26 & 16.67 & 17.03 & 16.67 & 15.76 & 16.61 & 17.18 & 16.68\\
                        ~& $v_x$  Avg / (m/s) & 15.72 & 15.81 & 15.93 & 15.84 & 15.98 & 15.97 & 16.05 & 15.98 & 13.13 & 12.56 & 16.19 & 16.12\\
                             ~& $\kappa$  Avg & 0.57 & 0.57 & 0.57 & 0.56 & 0.56 & 0.57 & 0.55 & 0.55 & 0.49 & 0.50 & 0.61 & 0.60\\
\hline
         \multirow{6}{*}{Competitor-3} & TTL Avg / (s) & 3.56 & 3.85 & 3.22 & 3.21 & 3.18 & 3.04 & 3.34 & 3.29 & 3.56 & 3.37 & 3.28 & 2.90\\
                             ~& LCT Avg / (s) & 6.35 & 6.19 & 4.92 & 5.18 & 5.21 & 5.00 & 5.74 & 5.54 & 5.66 & 5.63 & 5.49 & 5.00\\
                              ~& SY Avg / (m) & 0.98 & 1.00 & 0.96 & 1.00 & 0.99 & 1.00 & 1.00 & 1.00 & 1.35 & 1.00 & 1.12 & 1.00\\
                         ~& $v_x$ Max / (m/s) & 16.89 & 16.66 & 16.99 & 16.67 & 17.10 & 16.66 & 17.14 & 16.67 & 16.87 & 16.63 & 17.23 & 16.68\\
                        ~& $v_x$  Avg / (m/s) & 15.70 & 15.69 & 15.99 & 15.90 & 16.09 & 16.01 & 16.12 & 16.06 & 15.79 & 15.78 & 16.67 & 16.15\\
                             ~& $\kappa$  Avg & 0.55 & 0.54 & 0.60 & 0.60 & 0.54 & 0.51 & 0.60 & 0.61 & 0.57 & 0.56 & 0.55 & 0.54\\

			\hline\hline
		\end{tabular}
	}
\end{table*}

\begin{table*}[!t]
	\renewcommand{\arraystretch}{1.2}
	\caption{Lane-change trajectory similarity}
\setlength{\tabcolsep}{5 mm}
	\centering
	\label{table_6}
	%\centering
	\resizebox{\textwidth}{10mm}{
		\begin{tabular}{c | c c | c c | c c }
			\hline\hline \\[-4mm]
			\multicolumn{1}{c|}{} & \multicolumn{2}{c|}{Case 1} & \multicolumn{2}{c|}{Case 2} & \multicolumn{2}{c}{Case 3} \\
\cline{2-7} \multicolumn{1}{c|}{vs} & \makecell [c] {HD-A \& HLM-A} & \makecell [c] {HD-B \& HLM-B} & \makecell [c] {HD-A \& HLM-A} & \makecell [c] {HD-B \& HLM-B} & \makecell [c] {HD-A \& HLM-A} & \makecell [c] {HD-B \& HLM-B} \\
\hline
    \multicolumn{1}{l|}{Competitor-1} & 0.76 & 0.84 & 0.96 & 0.83 & 0.90 & 0.87 \\
    \multicolumn{1}{l|}{Competitor-2} & 1.00 & 0.80 & 0.92 & 1.00 & 0.68 & 0.84 \\
    \multicolumn{1}{l|}{Competitor-3} & 1.00 & 1.00 & 1.00 & 1.00 & 0.78 & 0.82 \\

			\hline\hline
		\end{tabular}
	}
\end{table*}

\begin{figure}[t]\centering
	\includegraphics[width=8.5cm]{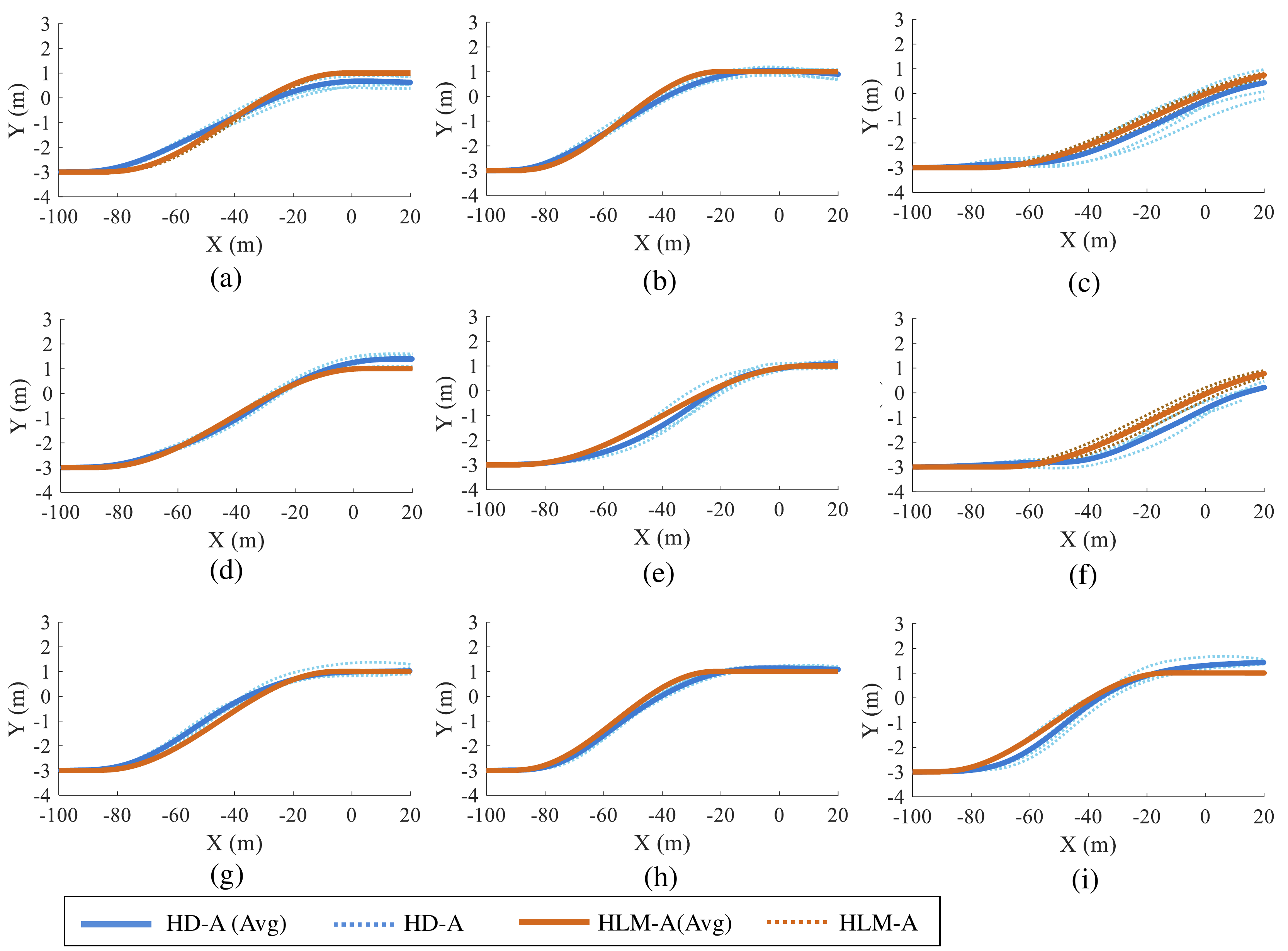}
	\caption{Lane-change trajectories of V1 considering interaction: (a) Competitor-1 vs HD-A/HLM-A in Case 1; (b) Competitor-1 vs HD-A/HLM-A in Case 2; (c) Competitor-1 vs HD-A/HLM-A in Case 3; (d) Competitor-2 vs HD-A/HLM-A in Case 1; (e) Competitor-2 vs HD-A/HLM-A in Case 2; (f) Competitor-2 vs HD-A/HLM-A in Case 3; (g) Competitor-3 vs HD-A/HLM-A in Case 1; (h) Competitor-3 vs HD-A/HLM-A in Case 2; (i) Competitor-3 vs HD-A/HLM-A in Case 3.}\label{FIG_13}
\end{figure}

\begin{figure}[t]\centering
	\includegraphics[width=8.5cm]{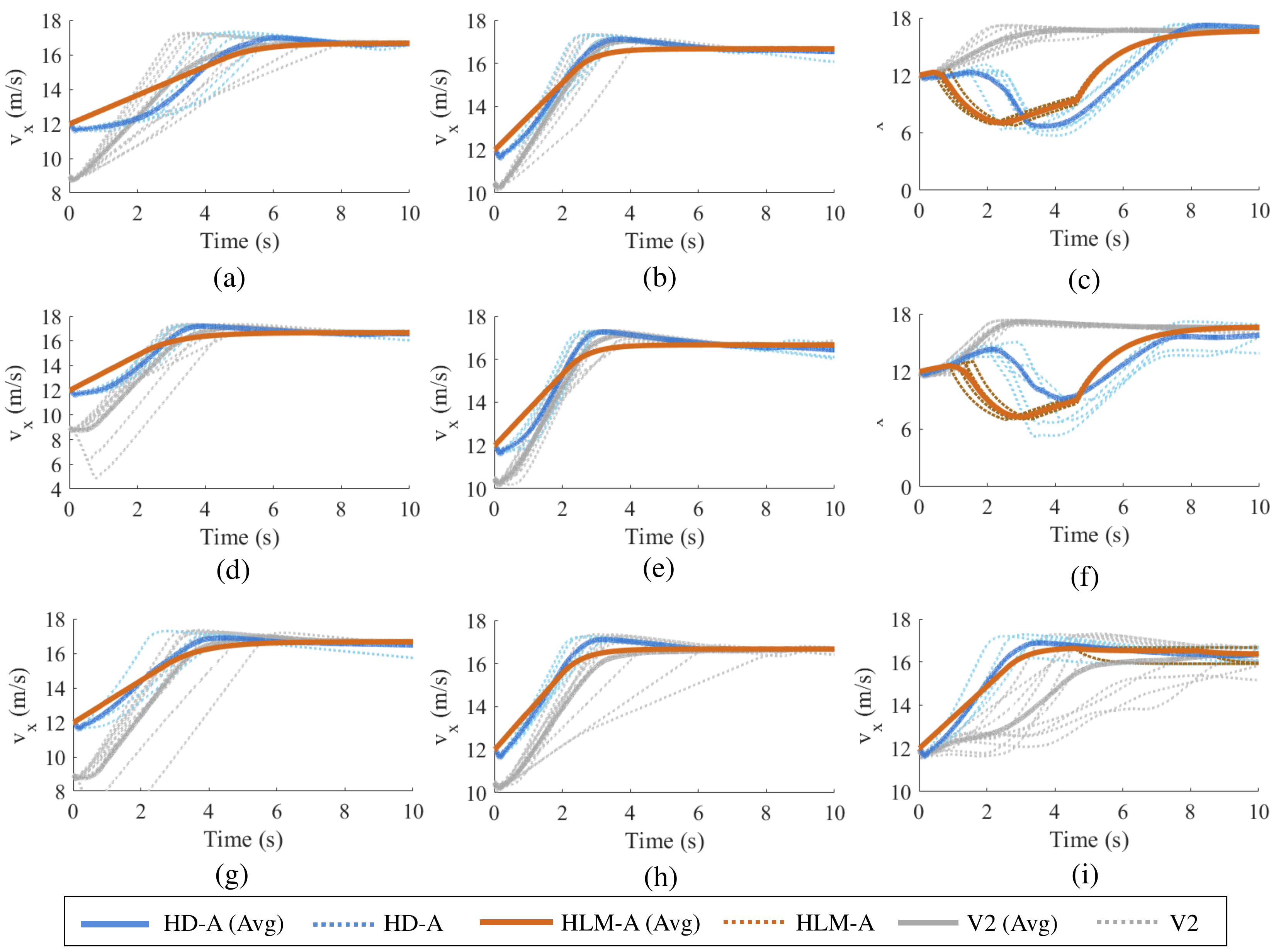}
	\caption{Lane-change velocities considering interaction: (a) Competitor-1 vs HD-A/HLM-A in Case 1; (b) Competitor-1 vs HD-A/HLM-A in Case 2; (c) Competitor-1 vs HD-A/HLM-A in Case 3; (d) Competitor-2 vs HD-A/HLM-A in Case 1; (e) Competitor-2 vs HD-A/HLM-A in Case 2; (f) Competitor-2 vs HD-A/HLM-A in Case 3; (g) Competitor-3 vs HD-A/HLM-A in Case 1; (h) Competitor-3 vs HD-A/HLM-A in Case 2; (i) Competitor-3 vs HD-A/HLM-A in Case 3.}\label{FIG_14}
\end{figure}

\begin{figure}[t]\centering
	\includegraphics[width=8.5cm]{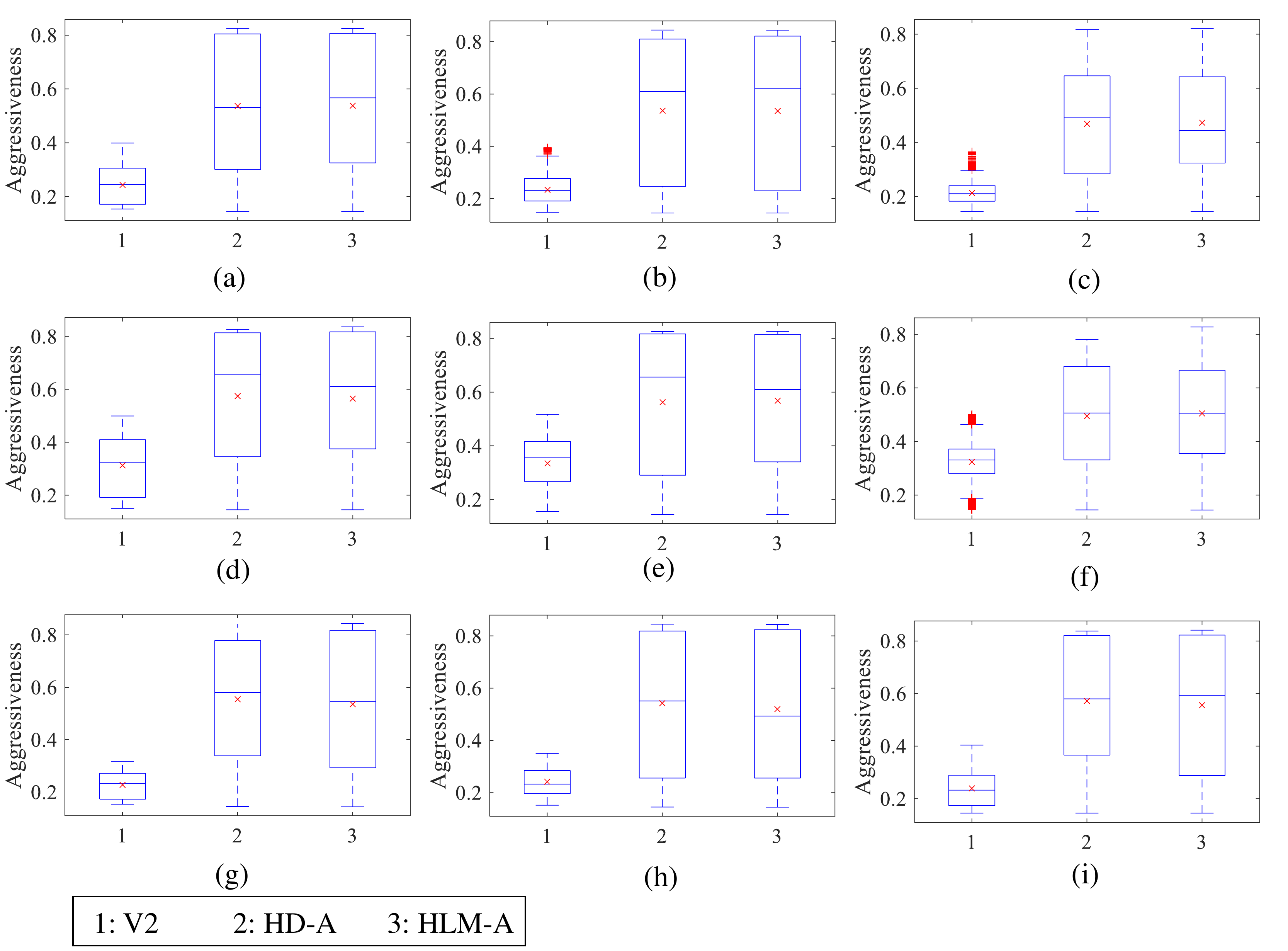}
	\caption{Aggressiveness of vehicles: (a) Competitor-1 vs HD-A/HLM-A in Case 1; (b) Competitor-1 vs HD-A/HLM-A in Case 2; (c) Competitor-1 vs HD-A/HLM-A in Case 3; (d) Competitor-2 vs HD-A/HLM-A in Case 1; (e) Competitor-2 vs HD-A/HLM-A in Case 2; (f) Competitor-2 vs HD-A/HLM-A in Case 3; (g) Competitor-3 vs HD-A/HLM-A in Case 1; (h) Competitor-3 vs HD-A/HLM-A in Case 2; (i) Competitor-3 vs HD-A/HLM-A in Case 3.}\label{FIG_15}
\end{figure}

In this experiment, V1 is controlled by human driver and human-like model to conduct the lane-change behavior. V2 is controlled by three human drivers, i.e.,  Competitor-1 (6 year driving experience), Competitor-2 (3 years driving experience), and Competitor-3 (1 years driving experience). V3 is set to be moving at constant speed. The initial position coordinates of V1 and V3 are set as (-100, -3) and (-65, -3), respectively. The initial velocities of V1 and V3 are set as 12 m/s and 5 m/s, respectively. The $Y$ coordinate position of V2 is 1. By controlling the relative distance and velocity between V1 and V2, it yields three cases. The initial $X$ coordinate of V2 is set as -125, -118 and -110 in Case 1, 2 and 3, respectively. Besides, the initial velocity of V2 is set as 9, 10.5 and 12 in Case 1, 2 and 3, respectively. Two human drivers (HD-A and HD-B), and two human-like models (HLM-A and HLM-B) are considered in the test. In each case, five repeated tests are conducted.

\begin{figure}[t]\centering
	\includegraphics[width=8.5cm]{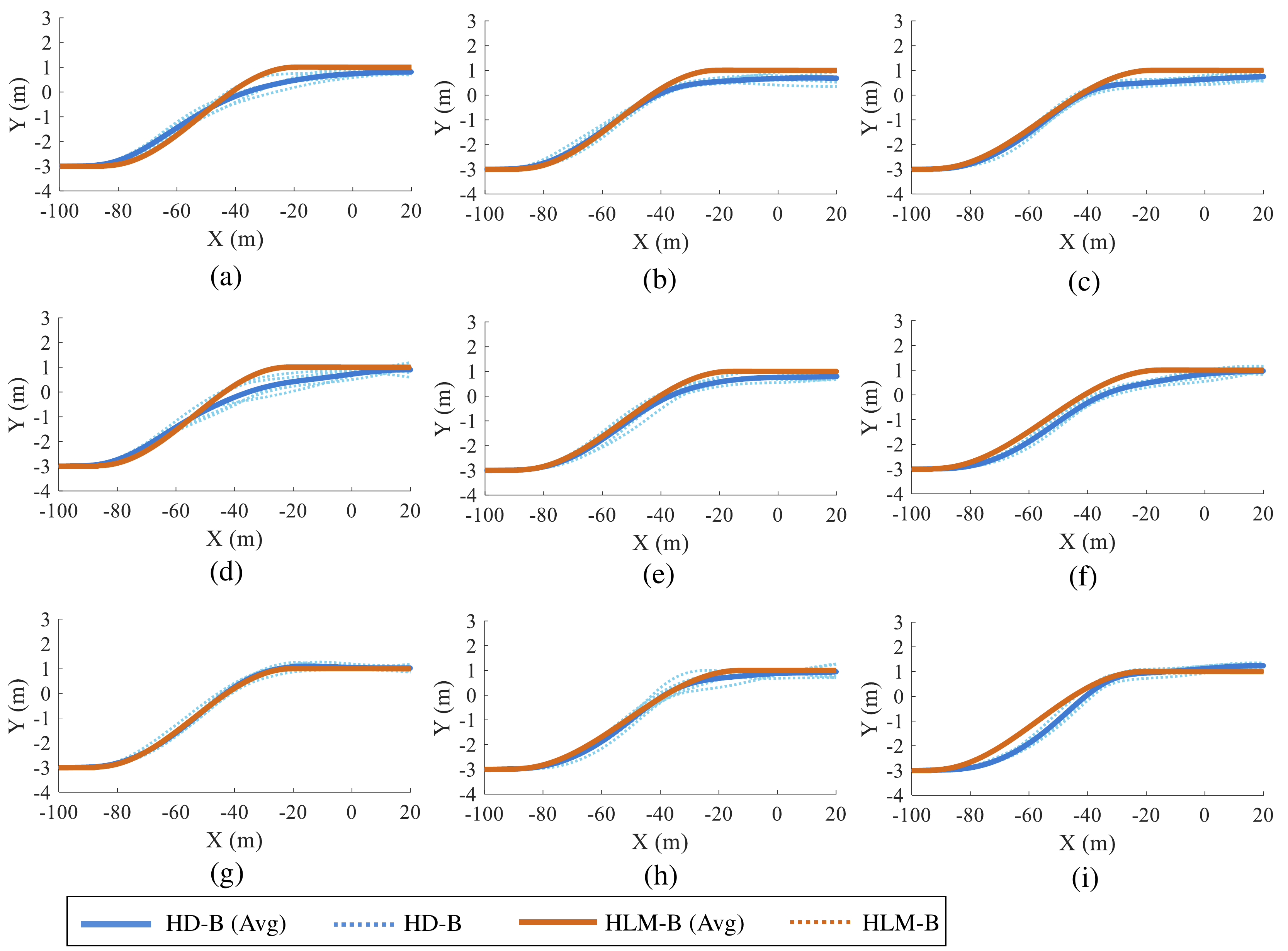}
	\caption{Lane-change trajectories of V1 considering interaction: (a) Competitor-1 vs HD-B/HLM-B in Case 1; (b) Competitor-1 vs HD-B/HLM-B in Case 2; (c) Competitor-1 vs HD-B/HLM-B in Case 3; (d) Competitor-2 vs HD-B/HLM-B in Case 1; (e) Competitor-2 vs HD-B/HLM-B in Case 2; (f) Competitor-2 vs HD-B/HLM-B in Case 3; (g) Competitor-3 vs HD-B/HLM-B in Case 1; (h) Competitor-3 vs HD-B/HLM-B in Case 2; (i) Competitor-3 vs HD-B/HLM-B in Case 3.}\label{FIG_16}
\end{figure}

\begin{figure}[t]\centering
	\includegraphics[width=8.5cm]{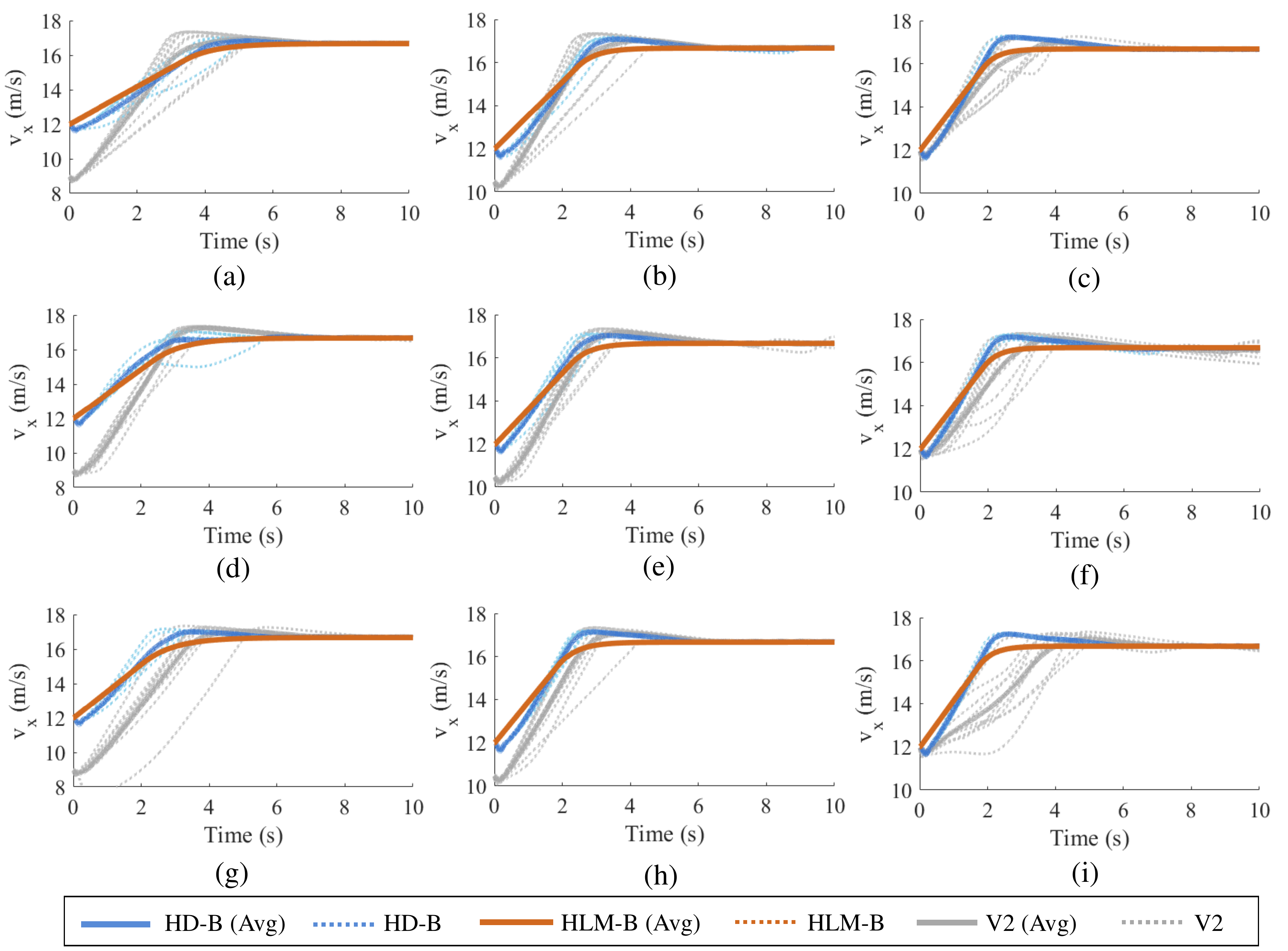}
	\caption{Lane-change velocities considering interaction: (a) Competitor-1 vs HD-B/HLM-B in Case 1; (b) Competitor-1 vs HD-B/HLM-B in Case 2; (c) Competitor-1 vs HD-B/HLM-B in Case 3; (d) Competitor-2 vs HD-B/HLM-B in Case 1; (e) Competitor-2 vs HD-B/HLM-B in Case 2; (f) Competitor-2 vs HD-B/HLM-B in Case 3; (g) Competitor-3 vs HD-B/HLM-B in Case 1; (h) Competitor-3 vs HD-B/HLM-B in Case 2; (i) Competitor-3 vs HD-B/HLM-B in Case 3.}\label{FIG_17}
\end{figure}

\begin{figure}[t]\centering
	\includegraphics[width=8.5cm]{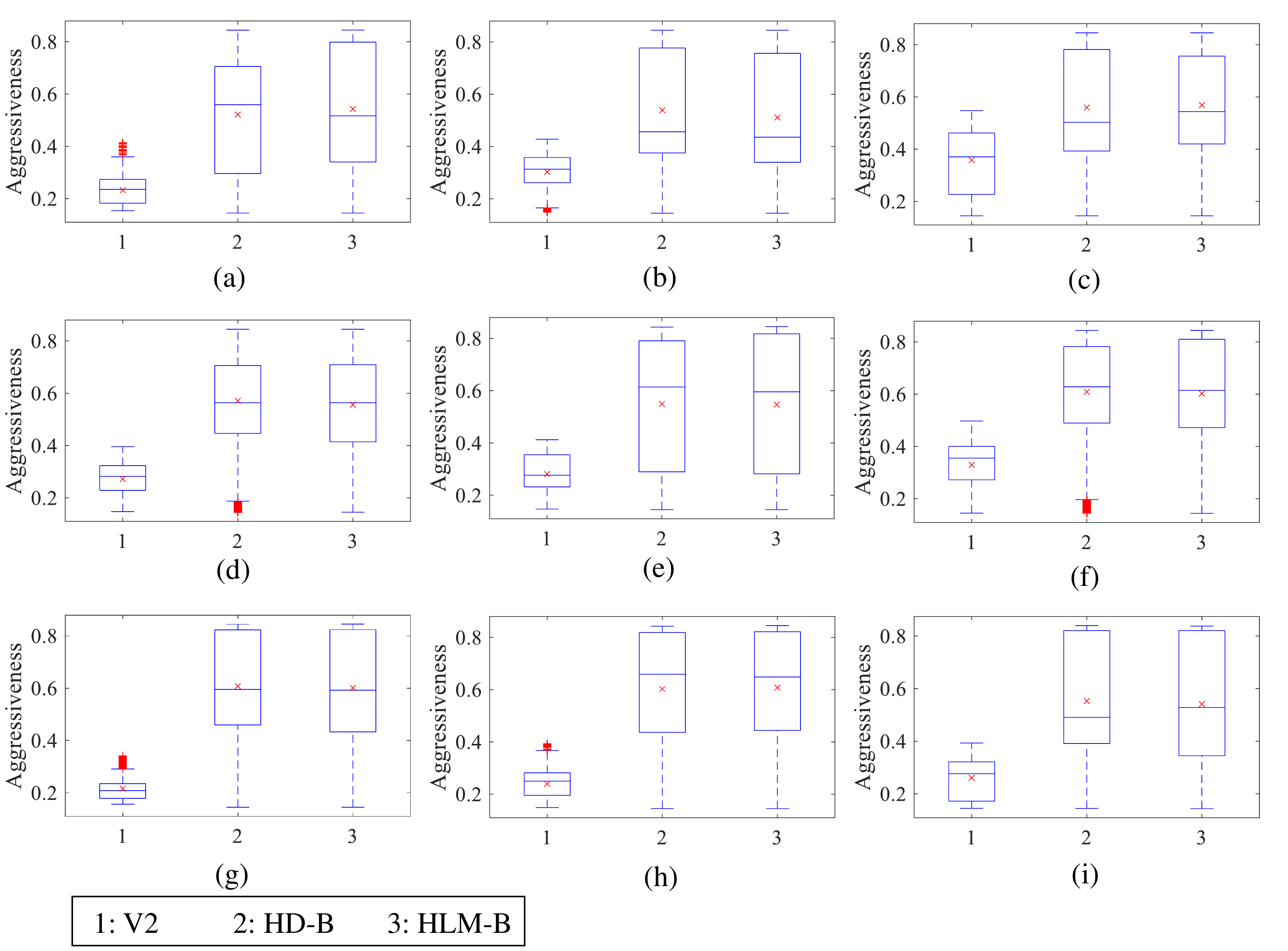}
	\caption{Aggressiveness of vehicles: (a) Competitor-1 vs HD-B/HLM-B in Case 1; (b) Competitor-1 vs HD-B/HLM-B in Case 2; (c) Competitor-1 vs HD-B/HLM-B in Case 3; (d) Competitor-2 vs HD-B/HLM-B in Case 1; (e) Competitor-2 vs HD-B/HLM-B in Case 2; (f) Competitor-2 vs HD-B/HLM-B in Case 3; (g) Competitor-3 vs HD-B/HLM-B in Case 1; (h) Competitor-3 vs HD-B/HLM-B in Case 2; (i) Competitor-3 vs HD-B/HLM-B in Case 3.}\label{FIG_18}
\end{figure}

The lane-change trajectories of V1 considering the interaction between HD-A/HLM-A and Competitor-1/Competitor-2/Competitor-3 are illustrated in Fig. 13. It can be found that the lane-change trajectories of HLM-A and HD-A are highly coincident in most cases, which verifies the human-like driving performances of the proposed algorithm. In most cases, both HD-A and HLM-A can change lanes easily. There exist some exceptions in the interaction between HD-A/HLM-A and Competitor-1/Competitor-2 in Case 3, i.e., Fig. 13 (c) and (f). Due to the small relative distance and velocity between V1 and V2, particularly the strong driving aggressiveness of Competitor-1 and Competitor-2, both HD-A and HLM-A chose give ways for Competitor-1 and Competitor-2 in Case 3. Besides, the results of lane-change velocities are illustrated in Fig. 14. In the interaction between HD-A/HLM-A and Competitor-1/Competitor-2 in Case 3, i.e., Fig. 14 (c) and (f), both HD-A and HLM-A chose decelerating firstly to guarantee the safe gap between V1 and V3. After V2 moved ahead, HD-A and HLM-A conducted the lane-change behaviors. Moreover, the velocity changes of HLM-A and HD-A are highly similar in most cases. Compared with HLM-A, HD-A has some speed drops at the beginning and some overshoots before reaching the steady state. The driving aggressiveness of drivers is estimated and displayed in Fig. 15. The aggressiveness distribution results of HD-A and HLM-A are highly similar in most cases, which means HLM-A has similar driving behavior to HD-A, verifying the human-likeness of the proposed algorithm.

The experimental results of the interaction and decision-making between HD-B/HLM-B and Competitor-1/Competitor-2/Competitor-3 are illustrated in Figs. 16, 17 and 18. The similar analysis results can be concluded, which will not be introduced repeatedly. It is worth mentioning that HD-B and HLM-B conducted the lane-change behaviors in all cases, which is different from HD-A and HLM-A in some extreme cases. The reason can be found from Fig. 18 that HD-B and HLM-B are more aggressive than HD-A and HLM-A.

The detailed test results are analyzed in Table V.
Besides, the Longest Common Sub-Sequence (LCSS) approach is used to evaluate the similarity of the lane-change trajectories. The LCSS of the two trajectories $A=\{a_1, \cdots, a_m\}$ and $B=\{b_1, \cdots, b_n\}$ is defined by\cite{2021A}
% Eq.
\begin{align}
LCSS(A, B)=
&
\left\{
\begin{array}{lr}
0,\quad if\quad  m=0 \quad or \quad n=0\\
1+LCSS(Rest(A),Rest(B)),\\
\quad \quad if \quad d(Head(A),Head(B))\leq\epsilon\\
\max
\left\{
\begin{array}{lr}
LCSS(Rest(A), B)\\
LCSS(A,Rest(B))\\
\end{array}
\right\},
 Other\\
\end{array}
\right.
\end{align}
where $Head(A)=a_1$, $Head(B)=b_1$, $Rest(A)=\{a_2, \cdots, a_m\}$, and $Rest(B)=\{b_2, \cdots, b_n\}$. Besides, $d(Head(A),Head(B))$ denotes the distance between $Head(A)$ and $Head(B)$, $\epsilon$ is the matching threshold.

Furthermore, the trajectory similarity ($TS$) is constructed as follows\cite{9419761}.
% Eq.
\begin{align}
TS(A,B)=\frac {LCSS(A,B)}{\min(m,n)}\in[0,1]
\end{align}

Based on Eq. 40, the analysis results of the lane-change trajectory similarity are displayed in Table VI. In most cases, the $TS$ is larger than $80\%$. The average $TS$ between HLM-A and HD-A is $89\%$, and the average $TS$ between HLM-B and HD-B is $88\%$, which indicates that both HLM-A and HLM-B are capable of conducting human-like lane-change behaviors.

According to the analysis results in Tables V and VI, some findings and conclusions are summarized as follows. Firstly, both the TTL and the LCT of V2 reduce with the decrease of the relative distance and velocity between V1 and V2, which means for guaranteeing driving safety, drivers usually tend to accelerate to finish the lane-change process. However, there exits an interaction and game process between the two drivers, and the results are affected by the driving characteristics, which is reflected by the driving aggressiveness in this paper. The interaction and game results between HD-A/HLM-A and Competitor-1/Competitor-2/Competitor-3, and between HD-B/HLM-B and Competitor-1/Competitor-2/Competitor-3 are quite different.

According to the analysis results of the key driving and decision-making indexes, i.e., TTL, LCT, SY, $v_x$, $\kappa$ and $TS$, it can be found that the designed human-like drivers (HLM-A and HLM-B) has similar driving and decision-making behaviors to human drivers (HD-A and HD-B). The human-likeness of the proposed driving and decision-making algorithm is verified.

The proposed human-like driving and decision-making algorithm is evaluated from three aspects, i.e., interacting with different opponents (Competitor-1, Competitor-2 and Competitor-3), designing different human-like drivers (HLM-A and HLM-B), and constructing different test cases (Case 1, Case 2 and Case 3). In general, the performance of the proposed human-like driving and decision-making algorithm has been validated from all aspects.

\section{Conclusion}
To deal with the interaction between AV and human drivers in the lane-change process, a human-like driving and decision-making framework is proposed in this paper. Based on the driving behavior analysis of human driver from the INTERACTION dataset, an aggressiveness estimation model is built with the fuzzy inference
approach. Besides, BELCM-based driving model is designed for AV's human-like driving. In the human-like lane-change decision-making algorithm, collision risk assessment is realized with the APF approach, which is used to guarantee the lane-change safety of AV. Both driving safety and travel efficiency are considered in the decision-making cost function. Furthermore, the dynamic game approach is applied to the interaction and decision making between AV and human driver. Finally, human-in-the-loop experiments are conducted to verify the performance the proposed algorithm. Experiment results indicate that the proposed algorithm is capable of realizing human-like driving and decision-making for AVs.

\ifCLASSOPTIONcaptionsoff
\newpage
\fi

\bibliography{Ref}

\end{document}